\documentclass[10pt]{article}

\usepackage{fullpage}
\usepackage{CJKutf8} 
\usepackage[utf8]{inputenc}
\usepackage{setspace}
\usepackage{parskip}
\usepackage{titlesec}
\usepackage[section]{placeins}
\usepackage{xcolor}
\usepackage{breakcites}
\usepackage{lineno}
\usepackage{hyphenat}

\PassOptionsToPackage{hyphens}{url}
\usepackage[colorlinks = true,
            linkcolor = blue,
            urlcolor  = blue,
            citecolor = blue,
            anchorcolor = blue,
            hypertexnames=false]{hyperref}
\usepackage{etoolbox}
\makeatletter
\patchcmd\@combinedblfloats{\box\@outputbox}{\unvbox\@outputbox}{}{}%
\makeatother

\usepackage[round,authoryear]{natbib}

\renewenvironment{abstract}
  {{\bfseries\noindent{\abstractname}\par\nobreak}\footnotesize}
  {\bigskip}

\titlespacing{\section}{0pt}{*3}{*1}
\titlespacing{\subsection}{0pt}{*2}{*0.5}
\titlespacing{\subsubsection}{0pt}{*1.5}{0pt}

\usepackage{authblk}

\usepackage{graphicx}
\usepackage[space]{grffile}
\usepackage{latexsym}
\usepackage{textcomp}
\usepackage{longtable}
\usepackage{tabulary}
\usepackage{booktabs,array,multirow}
\usepackage{amsfonts,amsmath,amssymb}
\usepackage{lettrine}
\newcommand{\dropcap}[1]{\lettrine[lines=2,lraise=0.0]{#1}{}}
\usepackage{tikz}
\usetikzlibrary{positioning,arrows.meta}
\tikzset{
  state/.style={circle,draw,line width=0.8pt,minimum size=9mm,inner sep=1pt,font=\small,align=center},
  node distance=1.8cm,
  every edge/.style={draw,->,line width=0.8pt},
  >=Latex
}
\providecommand\citet{\cite}
\providecommand\citep{\cite}

\newif\iflatexml\latexmlfalse

\AtBeginDocument{\DeclareGraphicsExtensions{.pdf,.PDF,.eps,.EPS,.png,.PNG,.tif,.TIF,.jpg,.JPG,.jpeg,.JPEG}}

\usepackage[utf8]{inputenc}
\usepackage[english]{babel}

\usepackage{float}
\usepackage{tcolorbox}

\NewTColorBox[auto counter]{widegraybox}{O{} O{}}{
    float*,
    floatplacement={!t},
    width=\textwidth,
    colback=gray!10,
    colframe=black!65,
    boxrule=0.5pt,
    arc=2pt,
    before skip=1em,
    after skip=1em,
    title={Box~\thetcbcounter:~#1},
    fonttitle=\strut\bfseries,
    coltitle=white,
    #2,
}

\begin{document}
\begin{CJK}{UTF8}{gbsn}

\title{Why is anything conscious?}

\author[1]{Michael Timothy Bennett\thanks{Correspondence: m@michaeltimothybennett.com}}
\author[2]{Sean Welsh}
\author[3,4]{Anna Ciaunica}

\affil[1]{School of Computing, Australian National University, ACT 2601, Australia}
\affil[2]{Engine No.2, Bardon, Brisbane, QLD 4605, Australia}
\affil[3]{Institute of Cognitive Neuroscience, University College London, WC1N 3AZ, London, UK}
\affil[4]{INESC-ID, Instituto Superior T\'ecnico, University of Lisbon, Lisbon, Portugal}

\vspace{-1em}

\date{February 12, 2026}

\begingroup
\let\center\flushleft
\let\endcenter\endflushleft
\maketitle
\endgroup

\vspace{0.5em}

\selectlanguage{english}
\begin{abstract}
We tackle the problem of consciousness by taking the naturally selected, embodied organism as our starting point. We provide a formalism describing how biological systems such as human bodies self-organize to hierarchically interpret unlabelled sensory information according to valence. The system is attracted and repelled at different spatial and temporal scales. This is a qualitative interpretation of an unlabelled physical state. We show how such interpretations imply behavioural policies which are differentiated from each other only by this qualitative aspect of information processing. Natural selection favours systems that actively intervene in the world to achieve homeostatic and reproductive goals. Put provocatively, death grounds meaning. This means that in living systems information processing is necessarily subjective, that is, it has quality embedded into its very core. Qualitative information processing involves interoceptive and exteroceptive classifiers, and determines priorities for self-survival. We formulate The Psychophysical Principle of Causality as a theorem, and prove generalisation optimal learning forces this valence first ontology. Qualitative good or bad processing necessarily comes \textit{before} quality neutral representations of properties (i.e. ``red’’ is constructed from valence). Under selection pressures like sophisticated predation this produces a hierarchy of selves, of which reafference and reflective self awareness are a consequence. We discuss this in light of the seminal distinction between phenomenal and access consciousness. We claim that phenomenal consciousness without access is likely common, but the reverse is implausible. Our proposal lays the foundation of a formal science of consciousness, closer to human fact than zombie fiction.

\end{abstract}%

\sloppy

\hypertarget{introduction}{%
\section{Introduction}}

\dropcap{W}hy is anything conscious? Both biological and other physical systems process information, yet humans consciously \textit{experience} as well as process the information. Why?
Living organisms are constantly processing self and world related information to survive in an ever-changing world. Human bodies share with all other physical systems the property of being instantiated in time and space (e.g. occupy space). Yet, unlike physical systems, biological systems are dissipative systems using energy to self-organise in the face of entropic decay and environmental perturbation \cite{nagel1974,chalmers1995}.

Initially formally introduced in the field of cybernetics \cite{ashby1947,foerster1960} the notion of self-organisation has been applied to various disciplines including physics \cite{haken1983}, biology \cite{camazine2003,bell2007} and neuroscience \cite{kelso1995,friston2010,tognoli2014}. Self-organisation is typically defined as the spontaneous emergence of spatiotemporal order or pattern-formation processes in physical and biological systems resulting from interactions of its components with the environment \cite{camazine2001,seeley2002,rosas2018}.

Interestingly, much self and world related information processing goes on behind the scenes, or ``in the dark'' so to speak, that is, without being constantly present to our conscious minds. But why doesn’t all information processing go in the dark? 

This question is the subject of long-standing debate \cite{seth2022}. One highly influential view is that consciousness has two aspects \cite{seth2022}. The first is functional, by which we mean the ability to \textbf{access} and communicate information \cite{block1995}. How function relates to consciousness is considered the ``easy problem'' of consciousness \cite{chalmers1995}. The second aspect is ``what it is like'' to consciously experience information processing, or \textbf{phenomenal} consciousness \cite{chalmers1995,block1995,nagel1974,gallagher2021,fuchs2017}. This doesn't mean just global states like being awake, but more local states like smelling a cup of coffee. These local contents or ``qualia'' are characterised by what it is like to be in them \cite{seth2022}. It is unclear the extent to which functional and phenomenal aspects are independent. David Chalmers has influentially suggested that it may be possible to construct a ``philosophical zombie'' which acts in every way like a person but has no qualia \cite{block1995}. For example, a thermostat certainly detects heat and so processes information, but there presumably is not anything it is like to be a thermostat. Hence the question ``why is anything conscious'' may be understood as ``why is there sometimes a qualitative aspect to information processing?''. This is somewhat related to the ``hard problem'' of consciousness. However the hard problem has sparked a substantial body of work, detailed discussion of which \cite{seth2022,northoff2014} lies beyond the scope of our paper. In what follows we largely avoid the mental/physical distinction that fuels the classical debates, ranging from panpsychism \cite{goff2019} to physicalism à la Dennett \cite{dennett1991}.

Likewise we remain agnostic to the question whether conceivability arguments such as the zombie thought experiment \cite{chalmers1996_consciousMindBook} or the knowledge argument \cite{jackson1982,jackson1986} are the best way to approach the problem of phenomenal consciousness.


How we do so can be understood in terms of the hardware and software metaphor. The distinction between mental and physical that fuels classical debates can be understood in computational terms. In this view mental activity is like software, which interacts with the physical world through hardware \cite{hutter2010,bennett2025b,orseau2012b}. Hardware acts as an interpreter between software and the physical environment. This is somewhat like the Pineal gland in Cartesian dualism, and so this view is called ``computational dualism'' \cite{bennett2024a}. While software is a useful frame for building standardised products, formalisms of intelligence based on computational dualism are inherently flawed \cite{bennett2024a,leike2015}. The problem is that software is not in fact a separate substance. In actuality, software is a state of the hardware on which it runs. To explain, computers can be understood as ``stacks'' of abstraction layers, where each layer is simplified subsets of behaviour of the layer below. For example Python code relies on interpretation by a program written in C, which is built on Assembly, which in turn relies on machine code and thus hardware. In this sense, software is an ``abstraction layer'' over the hardware on which it runs. Nature privileges adaptation, and working only at high levels of abstraction can limit adaptability, decreasing efficiency \cite{bennett2025thesis}. 
Compare the vast quantity of data and energy required to train a large language model to solve a problem, to the small quantities humans need to solve a problem. Biological systems are more efficient, because they delegate adaptation to lower levels of abstraction \cite{bennett2024c}. For example, an organism can be understood as an abstraction layer over organs, which are a layer over cells. Adaptation takes place at every level of this biological ``stack'', whereas a software ``intelligence'' in a computer is adaptive only at a a high level of abstraction, in software. Our ability to model and understand problems is limited by the level of abstraction at which we view them \cite{bennett2025thesis}. Efficiency problems in Artificial Intelligence (AI) can be addressed by accounting for abstraction layers \cite{bennett2024a}. Mental activity can be framed as a layer over the body, so rather than trying to explain the mind in the abstract like software, we start at the level of the embodied organism.

In short, our explanation of consciousness asserts that world has valence. Physical states are attractive or repulsive to a self-organising adaptive system. Organisms are such self-organising systems which learn to classify causes of valence that affect their decision procedures, to have a better chance of surviving. Qualia and affects, the atomic building blocks of phenomenal consciousness, are qualitative \textit{interpretations} rather than abstract logical or mathematical representations. Qualitative processing conveys an advantage in evolutionary terms. At high levels of abstraction these appear to be irreducible qualities to which valence seems attached. However at lower levels of abstraction these qualities are the various states of attraction and repulsion happening in a body at a given point in time. Put another way, sensory stimuli are attractive or repulsive at multiple spatial and temporal scales within a self-organising system \cite{bennett2025thesis}. Qualia and affect are easily and obviously linked to affordances in the world by organisms. Phenomenology has a `logic' – not the symbols and rules of the supersets of first order logic that make up programming languages but a presymbolic `logic' of cause and effect. These are modelled in the decision procedures of prelinguistic organisms as cause and affect. At a high level, organic decision procedures boil down to affect-infused qualia, the key components of phenomenology. Natural selection favours organisms with phenomenology. Not always but often. Our position is that phenomenology is functional and decisive, not epiphenomenal or illusory, but this is not to deny that conscious decisions have subconscious components. We see consciousness as the ``executive'' top level of a ``stack'' of biological components.

We bypass the ``computational dualism'' of hardware and software \cite{bennett2024a} by starting with low level biological facts rather than high level philosophical abstractions to reformulate the problem of consciousness in a way that does not take for granted the mental/physical division in the first place. 

Take for example the higher order thought theory (HOT) \cite{rosenthal2005,brown2019} holds that the information of which a conscious being is aware are higher order ``meta-representations'' of lower order ``local'' mental states. Lower order states may include emotions and perceptions, while higher order meta-representations reflect upon those. The link between the two may explain something of the phenomenal character of states. Sense data is processed by the body resulting in lower order mental states, and then meta representations of those is where we might find more abstract conceptual or thought-like contents of consciousness. 
HOTs might be a good starting point to understand why most biological information processing goes on ``in the dark'', but why do these lower order states arise? Can they occur in the absence of subjective, qualitative experience? Is there something it is like to ``be in the dark'' processing information at the lower bodily levels? 

Long considered a fringe approach, the embodied cognition paradigm \cite{varela2016} has recently gained substantial influence in cognitive science and philosophy \cite{ciaunica2023,levin2024,seth2018}. The idea is that instead of considering the body as a mere device designed to fuel and contain the mind (a device that can be replaced with a vat or a robot, for example), one must consider the mind as serving the self-sustaining needs of a surviving body. 

If this is so then unravelling consciousness must start with understanding the `humble' lower bodily levels of information processing, not the higher order, cognitive levels of information processing. In other words, conscious experiences do not merely depend on bodily experiences as an external factor that can be replaced with a vat or an artificial system. Rather they fundamentally constitute what a conscious experience is. How can we prove that?

Our aim here is to try to rigorously define consciousness from first principles and show that some aspects of functional consciousness depends on phenomenal consciousness in a manner that makes zombies impossible, ``dissolving'' the problem by showing the phenomenal to be intrinsically functional\footnote{Note that we aren't the first to claim that the phenomenal is functional. For example, the Conscious Turing Machine \cite{blum2020} based on Global Workspace Theory \cite{baars1997}, and constructivist approaches to AI \cite{wang2020} take similar positions. It is our explanation of how and why that is novel.}. We establish axioms that hold in every conceivable environment. We then argue that if those axioms hold there is no conceivable environment where the function of consciousness without the subjective experience of it, and why most information processing goes on ``in the dark''\footnote{That is, without explicit conscious awareness of it.}.
{\label{874460}}

\begin{widegraybox}[Supporting formal results and provenance][label=box:1]
This paper is primarily philosophical.
However, the key claims are also supported by formal theorems, proofs and experiments provided in the Supplementary Information (ESM).
Many of these experiments and proofs were peer reviewed and published in the PhD thesis ``How To Build Conscious Machines'' \cite{bennett2025thesis}, and in the proceedings of the Spring 2026 AAAI Workshop on Machine Consciousness \cite{bennett2026b}.
The main manuscript also introduces new formal results.
These include the proof of The Psychophysical Principle of Causality and substantially improved propositions showing how the 1ST, 2ND and 3RD order selves follow from generalisation optimal learning under the scale and incentive preconditions.
Full proofs are included in the ESM.
The reformulation of informal The Psychophysical Principle of Causality as a formal theorem, and the resulting proof, are dependent upon the already published theorem The Law of the Stack \cite{bennett2024c}, so The Law of the Stack is also detailed in the ESM.
Boxes 2 to 4 summarise the new theorem, the other proof results, and the experiments.
\end{widegraybox}

Enactivism is roughly the view that information processing arises through interaction between an organism and its environment \cite{thompson2007}, and it is considered by some to be incompatible with narrow definitions of computation. The notion of computation is widely debated, and a detailed review of these discussions would lead us to a major digression \cite{piccinini2015, piccinini2021}. Here we define computation not in terms of abstract symbol shuffling and representation, but in concrete mechanistic terms where `information processing' refers to the actual physical transition of a system from one state to another. To formalise enactivism\footnote{By formalising something akin to ``intelligence without representation'' as described by Brooks \cite{brooks1991}, we end up with physical interpretations rather than representations.}, we build upon Pancomputational Enactivism \cite{bennett2024a,bennett2024c} and Stack Theory \cite{bennett2025thesis}, which frame intelligence in terms of embodied abstraction layers.

To answer ``why is anything conscious'', we use this formalism to show how lower and higher order theories, phenomenal and access consciousness all follow from first principles, scaling natural selection pressures and the ability to adapt\footnote{This is loosely inspired by a scale-based framing of machine learning \cite{sutton2019} - we are applying Sutton's ``bitter lesson'' to biology.}. First, in Section \textbf{2} we integrate and extend separately published works of narrower scope \cite{bennett2024a,bennett2023b,bennett2023c} to justify our model of self-organising systems. In Sections \textbf{3} and \textbf{4} we explain how this model formalises relevance realisation and unifies lower and higher order theories. Rather than assuming abstract objects are primary and trying to learn causal relations between them, this assumes valence is primary and that abstract objects are constructed to classify cause and anticipate valence. This takes us from attraction and repulsion to physical states, through lower order thoughts to higher order meta representations. In the associated thesis \cite{bennett2025thesis} we call this the \textbf{psychophysical principle of causality}. Here we formalise this as a theorem of the same name (theorem 3 in the ESM). In Section \textbf{5} we extend previous work \cite{bennett2023c} on causal learning and the development of first (1ST), second (2ND) and third (3RD) order selves, and explain how they are a consequence of scaling the ability of a self-organising system to adapt with natural selection pressures. In Sections \textbf{6} and \textbf{7} we explain how subjective experience requires a 1ST order self, and conversely why a 1ST order self implies there is something it is like to be an organism that has a 1ST order self. We then argue that access consciousness requires the ability to communicate meaning, which requires both 1ST and 2ND order selves, and show how a philosophical zombie is impossible. In Section \textbf{8} we describe the development of consciousness as we scale up the capacity to adapt with natural selection pressures, with examples from nematodes to humans.

\section{Back to Foundations}
This paper can be read in purely narrative form, meaning the core argument should be understandable without the math. However for readers who do wish to delve into the mathematical details, a short summary of definitions is given below.
The full set of definitions used in this paper, along with full proofs and the full experimental protocols and plots, is in the Electronic Supplementary Material (ESM).
The longer Stack Theory appendix \cite{bennett2025thesis} is the master reference.
\begin{enumerate}
    \item An \textbf{environment} is a nonempty set $\Phi$ of mutually exclusive \textbf{states}. A (declarative) \textbf{program} is any set of states $p \subseteq \Phi$. Let $\mathcal{P} := 2^\Phi$ be the set of all programs. An \textbf{abstraction layer} is a finite \textbf{vocabulary} $\mathfrak{v} \subseteq \mathcal{P}$, meaning the finite set of programs a body can stably implement. A vocabulary induces an embodied \textbf{formal language}
    $$
        L_\mathfrak{v} = \{ l \subseteq \mathfrak{v} : \bigcap_{p \in l} p \neq \emptyset \}.
    $$

    \item Members of $L_\mathfrak{v}$ are called \textbf{statements}. The \textbf{truth set} of a statement $l$ is
    $$
        T(l) := \bigcap_{p\in l} p.
    $$
    We take $T(\emptyset)=\Phi$ by convention.
    The statement $l$ is true relative to a state $\phi \in \Phi$ iff $\phi \in T(l)$. A \textbf{completion} of $x$ is any statement $y$ such that $x \subseteq y$. The \textbf{extension} of a statement $x$ is the set of all its completions
    $$
        E_x := \{ y \in L_\mathfrak{v} : x \subseteq y\}.
    $$
    For a set $X \subseteq L_\mathfrak{v}$ we write $E_X := \bigcup_{x\in X} E_x$. Two statements are \textbf{equivalent} when they have the same extension.

    \item A $\mathfrak{v}$-task $\alpha$ is a pair $\langle I_\alpha, O_\alpha \rangle$ where $I_\alpha \subseteq L_\mathfrak{v}$ are the \textbf{inputs} and $O_\alpha \subseteq E_{I_\alpha}$ are the \textbf{correct outputs}. Tasks form a hierarchy. If $\alpha \sqsubset \omega$ then $\omega$ is a \textbf{parent} of $\alpha$, meaning $I_\alpha \subset I_\omega$ and $O_\alpha \subseteq O_\omega$.

    \item A \textbf{policy} is a statement $\pi \in L_\mathfrak{v}$. A policy is \textbf{correct} for $\alpha$ iff
    $$
        \pi \in {\Pi_\alpha} = \{\pi \in L_\mathfrak{v} : {E}_{I_\alpha} \cap E_{\pi} = {O_\alpha}\}.
    $$
    Given input $i \in I_\alpha$ and a policy $\pi$, to \textbf{infer} is to choose an output $o \in E_i \cap E_\pi$. A task is complete if $o \in O_\alpha$.

    \item Learning must choose among many correct policies. A standard proxy is \textbf{weakness} $w(\pi) := |E_\pi|$, which counts how many further commitments remain compatible with $\pi$ while staying correct. \textbf{Weakness maximisation} (w-maxing) chooses a correct policy that maximises $w(\pi)$. A \textbf{causal identity} is a policy $c^{\mathfrak{b}}_\mathfrak{a} \in L_\mathfrak{v}$ that classifies interventions by some source $\mathfrak{b}$ as distinct from matched observations, relative to a learner $\mathfrak{a}$. Nested causal identities represent higher orders of modelling. For example $c^{\mathfrak{ba}}_\mathfrak{a} \subseteq c^{\mathfrak{b}}_\mathfrak{a}$ is a refinement of $\mathfrak{a}$'s model of $\mathfrak{b}$ that depends on $\mathfrak{a}$'s prediction of how $\mathfrak{b}$ models $\mathfrak{a}$.
\end{enumerate}

\noindent We begin by formalising all conceivable environments \cite{bennett2024a,bennett2024b} from first principles:\\

\noindent \textbf{Axiom 1: \label{axiom1}} When there are things, we call these things the \textbf{environment}. \\

\noindent \textbf{Axiom 2: \label{axiom2}} Where things \textit{change} or \textit{differ}, we have different \textbf{states} of the environment. \\

Hence, the environment is a set $\Phi$ of mutually exclusive states. We call a set of states a declarative ``program''\footnote{We do not concern ourselves with imperative programs, because there is an equivalence between declarative and imperative \cite{howard1980}.}. A program is ``true'' relative to the states it contains. For example, if time is one dimension along which things change, then there is one state at a time. A ``\textbf{fact}'' is a program which contains the state at the present time. It follows that every aspect of the environment is a set of programs. An aspect of the environment is realised by (exists in) a state if the programs it contains are facts. Given each state is a difference, it follows that states are mutually exclusive, which elsewhere has been used to portray time as difference and each state as a point in time\footnote{Each possible sequence of states might be thought of as a timeline, and the set of all timelines can be thought of as the set of all conceivable universes (infinite if the set of states is infinite).}. Such foundational arguments are largely beyond the scope of this paper and are discussed at length elsewhere \cite{bennett2025thesis}. However, this frame allows us to talk about ``computation'' in a much more general sense than Turing computation. By computation, we mean physical cause and effect, dictated by the state transitions of whatever environment we might conceive of. In the frame of Stack Theory \cite{bennett2025thesis,bennett2024c}, this is like saying we don't know where the base abstraction layer is, whether there even is such a layer, or what it might entail. We are merely saying that no matter what abstraction layer is considered, there must be change or difference for there to be anything at all. That entails there must be states, which entails there are ontological ``facts'' or ``true programs'' we can formalise as sets of states, every aspect of the environment that exists is a set of such facts, and finally this lets us talk about computation in a much more general sense than Turing computation\footnote{Related work holds each timeline is a ``cosmic ought'' from which goals and value judgements follow \cite{bennett2025thesis}. From a subjective perspective of a body (an aspect of the environment), ``truth'' is unknown (there can be many possible timelines in which a given body exists, seeming non-deterministic), but from the objective perspective of a given timeline truth is deterministic. Deterministic and non-deterministic views can be reconciled, just as the set of all languages recognisable as DFAs (deterministic finite automata) is the same as that recognisable as NFAs (non-deterministic) \cite{rabinscott1959}. In this view, the universe has a sort of ``opinion'' in the sense that a given timeline preserves some aspects of the environment and destroys others as it progresses. The set of all possible timelines would be the set of all such ``opinions''.}. We can talk about any given environment, or any given physical body within that environment, as an abstraction layer that performs a unique set of computations\footnote{Put another way, no two bodies perform exactly the same set of computations.}.
To give some more intuition as to why we can talk about any body, or any conceivable environment in this frame, note that we assume nothing of what states might ontologically be or epistemologically contain. We assume no differences within a state, only between states, and those differences are the programs of which aspects are formed. After all, one can only interact with aspects of one's environment. If one were to try and pinpoint what an aspect was made of, the answer would be deferred to other aspects. Any description that sought to fully encapsulate the semantics, truth or unmediated pure experience of a state would always be delayed or incomplete. This does not render such attempts vacuous, but speaks to their inherent contingency and conditionality, which is what this formalism attempts to capture.

\subsection{Natural Selection and Embodiment}
We assume a naturally selected, embodied organism. Every aspect of the environment is a set of facts\footnote{Some may object, pointing out that this ignores composition. However, the application of a function is a fact regardless of whether it takes another function as input.}, hence the body of an organism that persists across multiple states must be a set of programs. We call this set a \textbf{vocabulary} (formally defined in the appendix \cite{bennett2025thesis} as well as the summary at the beginning of this chapter). 

The meaning of abstractions depends upon interpretation, much like the behaviour of software is determined by the abstraction layer that ``interprets'' it. At its least abstract software is machine code\footnote{A ``word'' of machine code triggers a mechanistic process hardwired in the CPU by the human design. For example a word may copy the 32 bit value stored at memory address $X$ into the 32 bit register $Y$, which the next line of machine code ``adds'' to register $Z$ by looping over each bit in an ``adder'' circuit.}. Machine code is interpreted by hardware which determines every aspect of what that machine code does. A word of machine code is a mechanical trigger that only ``means'' whatever we have designed the hardware to do when we input that word. That hardware is an abstraction layer in which the software exists (including ``higher level'' abstraction layers like the Python interpreter). As stated earlier, what we call software is nothing more than the state of hardware \cite{bennett2024a}. This has undermined all but the most subjective of claims regarding the behaviour of theorised, software superintelligence \cite{hutter2010,leike2015,leike2018}. It is a flaw in the very idea of intelligent software. The distinction between software mind and hardware embodiment is subsequently called computational dualism \cite{bennett2024a}, because it is reminiscent of how traditional Cartesianism conceived of a mental substance distinct from physical. 

If we start with the embodied biological organism instead of the abstract mental states, then we cannot presuppose organisms use particular abstractions. Instead we must explain why particular abstractions are formed, in terms of contentless states, as the product of interaction between an organism and its environment \cite{thompson2007,varela2016,rolla2021}. This process is sometimes called ``relevance realisation'' \cite{vervaeke2012,vervaeke2013a,vervaeke2013b}, and this shift in perspective is key for our argument here. Indeed, the idea that conscious experiences depend upon interactions with the environment has been defended by a longstanding tradition in embodied cognition and enactivism \cite{thompson2007}.

For example, some have argued enactivism is incompatible with computation at a fundamental physical level, because the set of possible abstractions that must be searched for relevance realisation is intractable \cite{jaeger2024}. For tractability, one must first isolate a ``small world'' of relevant information from the ``big world'' of all information \cite{savage1954}. To address this we formalise embodiment, in concrete terms from the level of irreducible physical states. Embodiment implies finite resource constraints \cite{bennett2022c,bennett2025thesis,ciaunica2025nobody}, hence we adopt one more axiom.\\

\noindent \textbf{Axiom 3: } All aspects of the environment are spatially extended \cite{bennett2024b}.\\

This means that a body is realised by only a finite number of states. See the ESM for the self contained definition, and the Stack Theory appendix \cite{bennett2025thesis} for the full development. A body is a \textbf{vocabulary} $\mathfrak{v} \subset \mathcal{P}$ with finitely many elements. $\mathfrak{v}$ implies a \textbf{formal language} $L_\mathfrak{v}$ of interaction between body and environment. A \textbf{statement} in this language is an aspect of the environment. A set of ``programs'' from our vocabulary. Statements have truth conditions with respect to environmental states, and every statement has an \textbf{extension}. The extension of a statement is the set of all statements which are supersets of the first statement (the set of all other statements by which the first statement is implied). Extension is important we can relate statements by their truth conditions, forming a lattice. Most importantly, we avoid the distinction between software and hardware and so avoid computational dualism.

To summarise. Embodiment is dispositional \cite{bennett2024a,bennett2025c}. As natural selection evolves a vocabulary, it begins to isolate a ``small world'' relevant to survival \cite{jaeger2024,bennett2024b,friston2023}, from the intractable ``big world'' $\mathcal{P}$ of all information. 

\subsection{Self-Organising Systems as Self and World Constraints}
Both snowflakes and human bodies self-organise, yet we regard only the latter as conscious. What separates the two? To tackle this question we need self and world constraints. 

Using our embodied formal language $L_\mathfrak{v}$, we can talk about computation with inputs and outputs by treating everything as embodied statements embedded and enacted within the environment. 
Given an \textbf{input} $i \in L_\mathfrak{v}$, the set of all possible \textbf{outputs} is the extension $E_i$ of that input. This is because if $i$ is realised by the environment, then the environment is constrained to those states that realise $i$, which constrains what other statements are realised. If behaviour is ``motivated'', then statements have \textbf{valence} determined by natural selection and only a subset of the statements a body could make are ``fit''. Hence an organism will self-organise to express some statements, but not others. Crucially, though environmental states are contentless they imply content by how they relate to one another. As states transition from one to another this preserves some aspects of the environment, and deletes others (which would happen when an aspect is made of facts that don't hold at a particular moment in time, given the state at that time). As argued elsewhere, the environment preserves that which preserves itself \cite{bennett2025thesis}. It follows that natural selection must make states attractive or repulsive to a self-organising system if that self-organising system is to be deemed `fit'. This is formalised by the $\mathfrak{v}$-task as in definition 3\footnote{A $\mathfrak{v}$-task is called a $\mathfrak{v}$-task because it is defined relative to a vocabulary $\mathfrak{v}$.}. If $\mathfrak{v}$ is the vocabulary of a body, and $\mathfrak{v}$-task $\mu = \langle I_\mu, O_\mu \rangle$ is fit behaviour, then $I_\mu$ is all the statements that body can express in which it is possible to remain fit, and $O_\mu$ is all the statements that body can express in which it remains fit. The extension $E_{I_\mu}$ of $I_\mu$ would be every output that it is \textit{possible} to choose given the inputs $I_\mu$, but only a subset of those $O_\mu \subset E_{I_\mu}$ are \textbf{correct outputs}.  

\subsection{Inference}
Every statement in $L_\mathfrak{v}$ implies a constraint, because there are only so many outputs that can be expressed by a body at the same time as any given statement. Assume we want to constrain the body to $O_\mu$ given $I_\mu$. A body could express a statement $\pi$ (meaning $\pi$ is true), and $\pi$ could then constrain the body to only correct outputs $O_\mu \subset E_{I_\mu}$ if $O_\mu = E_{I_\mu} \cap E_\pi$.
We call a constraining statement a \textbf{policy}. A policy constrains outputs given inputs, and a \textbf{correct policy} is one that constrains outputs to only \textbf{correct outputs}. For the sake of intuition, think of ``correct'' as ``fit'' according to natural selection\footnote{This need not always be the case.}. This is formalised in definition 4.

\subsection{Learning}
If an organism remains alive, then its history is an ostensive definition of ``fit'' self-organising behaviour. That history is expressed as a $\mathfrak{v}$-task $\mathfrak{h}$, where each input and output in $\mathfrak{h}$ is an interaction between organism and environment. If the set of all fit behaviour is a task $\mu$, and the organism remains fit throughout its history, then $\mathfrak{h} \sqsubset \mu$.

\begin{quote}
    \begin{center}
        \textit{
``I survive therefore my model is viable.'' - Mark Solms \cite{solms2021}}
    \end{center}
\end{quote}

$\mathfrak{h}$ implies a set $\Pi_\mathfrak{h}$ of policies, and not all of those policies will constrain outputs the same way given new inputs. Some policies will ``generalise'' to imply fit behaviour in unfamiliar circumstances, meaning correct outputs given a new set of inputs. If those policies imply fit behaviour given new inputs, then the organism will remain fit. An organism learns fit behaviour by learning $\pi \in \Pi_\mathfrak{h}$ such that $\pi \in \Pi_\mu$. 

\begin{quote}
    \begin{center}
        \textit{
``The best model of the world is the world itself'' - Rodney Brooks \cite{dreyfus2007}}\end{center}
\end{quote}

Earlier work \cite{bennett2023b} showed that the optimal strategy to adapt as fast as possible\footnote{Meaning to converge on a fit policy given the smallest history possible, to realise fit behaviour.} is to prefer ``weaker'' policies, meaning those with larger extensions. The policy which generated outputs $O$ given inputs $I$ is most efficiently identified by constructing a policy $\pi$ such that $\pi$ generates $O$ from $I$, and $\pi$ implies the weakest constraint that can be implied while still generating $O$ from $I$. This means a system that does so will construct fit policies from a shorter history \cite{bennett2023c} than one that does not. Full proofs are in the ESM.\footnote{The effect of insufficiently weak interpretations can be observed in large language models like GPT-3 and GPT-4 \cite{floridi2020,bennett2023d}, where they can fail to make consistent statements about an object because they represent it as multiple objects.}

\section{Relevance Realisation Through Causal Learning}
When an embodied organism learns a policy, it learns a relevant interpretation of that information, labelling it through behaviour.
To efficiently adapt, an organism must correctly anticipate valence, which means correctly identifying what causes valence. 

All organisms must have preferences, meaning some futures are better for them than others.
Formally, we model an organism as an embodied learner defined by an embodied vocabulary, a viability task, and a preference ordering over tasks. The full tuple definition is in the ESM.
Behaviour implies policies, and every policy implies tasks, so preferences can be formalised as a binary relation over tasks.

Natural selection favours adaptability, hence it prefers organisms that optimise for weak \textit{correct} policies, which is referred to elsewhere as \textbf{weakness maximisation} (w-maxing) \cite{bennett2025b}. By constructing fit policies, an organism divides the world into relevant objects and properties. A weaker policy implies all more specific versions of itself, meaning those that more tightly constrain outputs by having a smaller extension. Hence policy learning implies a lattice of policies that vary in weakness. Every fit policy would be a ``causal identity'' for something relevant, for example a specific object like ``this coffee'' or a weaker and more general concept like ``all coffee''. 

Organisms interact when they can affect one another in the causal or physical rather than psychological sense of the word.
Here affect means one organism's interventions can change the other organism's inputs, hence changing which outputs are viable. The formal definition we use is in the ESM.
Now, AI and machine learning \cite{russell2005ai,bishop2006,sutton2018,goertzel2014} are concerned with engineering adaptive agents. In that context, causality has become a mainstream topic of research. Causal learning is demonstrably necessary to thrive in an interactive setting \cite{pearl2018,richens2024robust}.
Where the causal graph is known in advance (for example if measuring the efficacy of medical interventions), this issue could be resolved using causal language \cite{pearl2009} to represent an intervention \textit{from outside} the system the graph describes.

To illustrate \cite{bennett2023c}, suppose Bob is attempting to learn and predict the environment. Bob has observed Alice wearing a raincoat only when it rains, and Bob has seen rain only when he observed Alice wearing a raincoat.
If we represent the raincoat observation with a binary variable $O \in \{true, false\}$ and the advent of rain with a variable  $R \in \{true, false\}$, and if Bob is a Bayesian, then Bob's observations will lead to the conclusion that $p(R=true \mid O=true) = 1$. This means Bob believes that if Alice wears a raincoat, then it must be raining. 
Now let us permit Bob to interact with his environment. Assume Bob wants it to rain. As $p(R=true \mid O=true) = 1$, Bob may conclude that forcing $O=true$ by holding a gun to Alice's head and demanding Alice wear a raincoat will cause it to rain. This is absurd. $O=true$ does not represent the event $q = $``I coerced Alice into wearing a raincoat'', but an entirely different event $v = $``Alice decided to put on a raincoat for the same reason I have observed Alice wearing a raincoat in the past''. To accurately represent the environment, we need to represent that $p(R=true \mid q) = p(R=true) \neq p(R=true \mid v) = 1$, meaning both $q$ and $v$. To illustrate visually, we started with the acyclic graph $$
\begin{tikzpicture}
    \node[state] (o) at (0,0) {$O$};
    \node[state] (r) [right =of o] {$R$};
    \path (r) edge (o);
\end{tikzpicture}
$$
and our intervention disconnected rain from the choice of clothing:
$$
\begin{tikzpicture}
    \node[state] (o) at (0,0) {$O$};
    \node[state] (r) [right =of o] {$R$};
\end{tikzpicture}
$$
This can be resolved by introducing a ``do'' operator that we apply to a variable $O$ to obtain $do[O=o]$, to represent the fact that an agency from outside the system has intervened to assign a value to $O$, so that we can represent $p(R=true \mid do[O=true]) = p(R=true) \neq p(R=true \mid O=true) = 1$. Thus, the aforementioned $q$ is equivalent to $do[O=true]$, while $v$ is equivalent to $O=true$.

\subsection{The Psychophysical Principle of Causality}
We have established the \textit{need} for causal reasoning, but not how one might come to know the objects involved (where did the variables come from?), or how they relate to one another causally (to form a graph). 

There are two ways to learn causal graphs. One can either assume a set of variables and learn relations between them, or assume relations and learn the objects that fit them \cite{bennett2023c}. We do the latter. Rather than assuming objects are primary and trying to learn causal relations between them, this assumes valence is primary and that abstract objects are constructed according to what \textbf{causes} valence. We call this the \textbf{psychophysical principle of causality}. This is what will take us from attraction to and repulsion from contentless physical states, through lower order thoughts to higher order meta representations. We start with the complete absence of content \cite{thompson2016}, and once a system learns policies there are distinct ``contents''. Policies serve to classify objects and properties. Death grounds meaning.

To develop this idea, we first show adding additional nodes to a causal graph is an adequate substitute for the ``do'' operator \cite{dawid2002}. Assume we again have Bob who constructs a causal graph of the environment. 
Assume Alice exists in that environment.
From Bob's perspective, Alice is just a part of the environment represented by a variable $A$ in Bob's causal graph.
$$
\begin{tikzpicture}
    \node[state] (a) at (0,0) {$O$};
    \node[state] (o) [right =of a] {$A$};
    \node[state] (r) [right =of o] {$R$};
    \path (o) edge (a);
    \path (r) edge (o);
\end{tikzpicture}
$$
Now assume Bob observes Alice taking an action that changes an aspect of the environment, represented by the variable $O$ (for example, Bob observes Alice putting on the raincoat). 
From Bob's perspective, Alice's action assigns values to variables $A$ and $O$.
There is no need to involve a do operator in this scenario because we can already represent that $p(R=true \mid A=x,O=true) = p(R=true) \neq p(R=true \mid A=y,O=true) = 1$ (because Alice is part of the causal graph).
Likewise, we can add variables for interventions of Bob \cite{bennett2023c}. 
$$
\begin{tikzpicture}
    \node[state] (i) at (0,0) {$O$};
    \node[state] (v) [above =of i] {$B$};
    \node[state] (o) [right =of i] {$A$};
    \node[state] (r) [right =of o] {$R$};
    \path (v) edge (i);
    \path (o) edge (i);
    \path (r) edge (o);
\end{tikzpicture}
$$

Now we will show how it is possible to \textit{learn} these nodes. The matter of \textit{which} cause-and-effect relations are learned is determined by valence, and so the nodes learned are statements classifying \textit{causes} and \textit{valence}.

\section{Relevant Causal Identities}
The optimal choice of policy for adaptability is the weakest because it is implied by more statements in an embodied language. That language is dispositional \cite{bennettmaruyama2022a}, shaped by natural selection, and so a weaker policy is likely to be a fit policy. w-maxing will correctly identify cause and effect where relevant, simplifying the entire environment into objects and properties relevant to the organism's motivations. 

Assume a vocabulary $\mathfrak{v}_\mathfrak{b}$ belonging to an organism $\mathfrak{b}$ (Bob). 
A ``cause'' in the context of this formalisation is not a variable but a \textit{statement} $l \in L_{\mathfrak{v}_\mathfrak{b}}$. The raincoat example would involve $obs, rain \in L_{\mathfrak{v}_\mathfrak{b}}$ such that:
$$obs \leftrightarrow \textit{``Alice put on a raincoat''} \text{ and } rain \leftrightarrow \textit{``It rained''}$$
$obs$ and $rain$ have truth values in accord with the definition. As in the earlier example Bob's passive observation implies $p(rain \mid obs) = 1$. The statement $obs = \textit{``Alice put on a raincoat''}$ can be made true by either passive observation or intervention. However, the statement which is true in the case of intervention not \textit{only} $obs$, but $int \in L$ such that $obs \subseteq int$ and: $$int \leftrightarrow \textit{``Alice is wearing a raincoat because of Bob's actions''}$$ 
$$
\begin{tikzpicture}
    \node[state] (i) at (0,0) {$\mathrm{int}$};
    \node[state] (o) [right =of i] {$\mathrm{obs}$};
    \node[state] (r) [right =of o] {$\mathrm{rain}$};
    \path (i) edge (o);
    \path (r) edge (o);
\end{tikzpicture}
$$
So long as $obs \neq int$, the intervention \textit{can} be differentiated from the passive observation at the more abstract representational scale of the vocabulary \cite{bennett2024c}.

This being the case, any set $c \subseteq int - obs$ could be used to identify the party undertaking the intervention, which is why $c$ is referred to as a ``causal identity''. It distinguishes the intervention $int$ from the passively observed effect $obs$, like reafference in living organisms. However, the above only considers one intervention. A \textit{weaker} or more general causal identity would be one that is shared by more interventions. 


A formal proposition shows that, when intervention episodes share at least one distinguishing signature, a w-max learner selects an inclusion minimal causal identity.
In plain English, it learns the least committal intervention tag that still separates intervention from matched observation.
See Box 3 for the sketch and see the ESM propositions on causal identity minimality for the full proof.

If a statement $l$ holds in both observation and intervention episodes, then $l$ alone cannot distinguish them.
If $c\in\mathcal{C}(INT,OBS)$, then $l\cup c$ is true on every $int\in INT$ and false on every $obs\in OBS$.
So $l\cup c$ plays the role of a do tag at the abstraction scale of $L_{\mathfrak{v}_\mathfrak{a}}$.

For example, suppose the inputs Alice has been subject to are $I_{\mathfrak{h}_{<t_\mathfrak{a}}}$. These can be divided into those in which Bob affected Alice $I^\mathfrak{b}_\mathfrak{a}$ and those in which Bob did not $I^{\lnot\mathfrak{b}}_\mathfrak{a} = I_{\mathfrak{h}_{<t_\mathfrak{a}}} - I^{\mathfrak{b}}_\mathfrak{a}$. Alice can construct a causal identity $b$ for Bob corresponding to interventions $INT = I^\mathfrak{b}_\mathfrak{a}$ and observations $OBS = I^{\lnot\mathfrak{b}}_\mathfrak{a}$. The full formalism for building these sets from a history is in the ESM.

\subsection{Ascribing Intent to Other Objects}
The distinction between ``intervention'' and not is misleading. Passive observation is just bearing witness to an intervention by something other than one's self. The question is not ``is this an intervention" but ``by whom was this intervention made?". 

Earlier, we arrived at the following graph in which Bob's intervention was given by $int$.

$$
\begin{tikzpicture}
    \node[state] (i) at (0,0) {$\mathrm{int}$};
    \node[state] (o) [right =of i] {$\mathrm{obs}$};
    \node[state] (r) [right =of o] {$\mathrm{rain}$};
    \path (i) edge (o);
    \path (r) edge (o);
\end{tikzpicture}
$$What if a third person Larry puts the coat on Alice? Bob may observe this, and so Bob's observation of Larry's intervention is $v \in L_{\mathfrak{v}_\mathfrak{a}}$ such that $obs \subset v$. To account for this, Bob can construct a causal graph as below (with $b.$ representing Bob and $l.$ representing Larry).

$$
\begin{tikzpicture}
    \node[state] (i) at (0,0) {$b.\,\mathrm{int}$};
    \node[state] (v) [above =of i] {$l.\,\mathrm{int}$};
    \node[state] (o) [right =of i] {$\mathrm{obs}$};
    \node[state] (r) [right =of o] {$\mathrm{rain}$};
    \path (v) edge (o);
    \path (i) edge (o);
    \path (r) edge (o);
\end{tikzpicture}
$$Bob's causal identity for himself $c_b \subset b. \ int - obs$ only represents the intervention by himself. However, now we can see that Bob must also construct an identity $c_l$ for Larry, where the  $c_l \subset l. \ int  - obs$. For an organism $\mathfrak{a}$ with an embodied language $L_{\mathfrak{v}_\mathfrak{a}}$ to construct a causal identity for an object $\mathfrak{b}$, it must first be the case that $\mathfrak{a}$ is affected by $\mathfrak{b}$ \cite{bennett2023d}, to satisfy the \textit{incentive} precondition for causal identity. \\
Assume an organism $\mathfrak{a}$ is affected by $\mathfrak{b}$ given inputs $INT$, and not affected given inputs $OBS$.
To then attribute the contents of $INT$ to one specific entity, there must be something in common between the members of $INT$ caused by $\mathfrak{b}$ that is not shared by any member of $OBS$ caused by something else (in other words it must be at least possible for $\mathfrak{a}$ to discern the existence of $\mathfrak{b}$). 
The contents of $INT$ are ``interventions'' by $\mathfrak{b}$ and by learning $c$, a corresponding causal identity, $\mathfrak{a}$ can discern the existence of $\mathfrak{b}$. This is not to say that $\mathfrak{b}$ \textit{has} intent, but intent can be ascribed to $\mathfrak{b}$ because $\mathfrak{b}$ affects $\mathfrak{a}$, who can then discern when interventions are a consequence of $\mathfrak{b}$.

A likelihood ratio bound shows that decoder mismatch can hide minds.
If every mind-like hypothesis in an evaluator's model fits an observed trace worse than a noise hypothesis by a large factor, Bayes rule forces the posterior away from mind (see Box 3 and ESM for technical details).

\subsection{Preconditions}
There are preconditions for the existence of a causal identity. 
First, the vocabulary $\mathfrak{v}_\mathfrak{a}$ of an organism $\mathfrak{a}$ must be of sufficient \textbf{scale} to ensure that observations are \textit{distinguishable} from interventions.
Second, there must be an \textbf{incentive} to construct it. Inference is only possible if some states are preferable to others. One cannot derive what ``ought'' to be from a statement of what ``is''. Natural selection provides a notion of what ought to be, by eliminating anything which ought not. In a creature that can die, death grounds meaning \cite{bennett2025thesis}.

\begin{enumerate}\label{preconditions_ci}
    \item The \textbf{scale} precondition requires $\mathfrak{v}$ contain the causal identity. 
    \item The \textbf{incentive} precondition is that fitness \textit{demands} the causal identity.
\end{enumerate}

\subsection{Realising Lower Order States And Higher Order Meta Representations}

Each policy an organism $\mathfrak{o}$ learns implies $\mathfrak{v}$-tasks. A $\mathfrak{v}$-task is a triadic relation between inputs, outputs and policies. This resembles Peircean semiosis \cite{sep-peirce-semiotics,bennett2023d} of sign, referent and interpretant. We call the resulting family of reusable tasks a ``protosymbol''\footnote{Proto because it is more primitive than a Peircean symbol.} system $\mathfrak{s}_\mathfrak{o}$ for the organism $\mathfrak{o}$. Intuitively, a protosymbol is a behavioural template that can be reused across contexts (full definition in ESM). Tasks exist in a ``generational hierarchy''. They are not mutually exclusive. Higher level tasks are more general, and have fewer policies because only very weak policies could complete them. Related as they are in a lattice, protosymbols are analogous \textbf{lower} order states and \textbf{higher} order meta representations. Some have framed consciousness as a problem of moving from unary, to dyadic, to triadic relations \cite{goertzel2006}. Similarly, we have gone from unary states to dyadic programs, to triadic tasks and protosymbols.

\begin{widegraybox}[Theorem 3. The Psychophysical Principle of Causality.][label=box:valence_first_thm]

One can learn causality by learning the objects and properties that fit a causal relation, or one can learn causal relations that fit a set of presupposed objects and properties. However if you want generalisation optimal learning, you cannot hard code a quality neutral ontology (no presupposed objects and properties).
Only the good or bad signal can be a fixed representational primitive from which more abstract classifications are constructed as needed. This is because valence is the signal that determines how representations are judged, and so any fixed abstraction which is not a direct consequence of that signal only serves to obfuscate the signal and impose a floor on variational free energy \cite{bennett2024c}. Continued existence and its lack determine what is attractive to whatever remains. Hence causal relation of valence is given. In a living system a causal identity is constructed of valence. It is a pattern in valence. A sufficiently complex mixture of repulsive and attractive signals, can together amount to a qualitatively neutral classification like ``the colour red''. Attraction and repulsion is how an organism carves up its world into objects and properties. Objects and properties are made of valence, and so are qualitative rather than quantitative information. This is the informal version of The Psychophysical Principle of Causality \cite{bennett2025thesis}. Here we reformulate this idea as an equivalent mathematical theorem, and prove (rather than just argue) that optimality demands valence first learning (at least, within the formal confines of the Stack Theory). We number this as Theorem 3 to match the numbering in the Supplementary Information.
Theorem 3 is a corollary to The Law of the Stack (see ESM), but is itself a novel contribution of this paper.

\medskip
\textbf{Definition (hard primitive commitments).}
Let $\alpha$ be a task and let $\Pi_\alpha$ be its set of correct policies.
Let $C\subseteq L_\mathfrak{v}$ be a nonempty set of commitments.
The restricted policy class that bakes in $C$ is
\[
\Pi_\alpha^C := \{ \pi\in \Pi_\alpha : \forall c\in C, c\subseteq \pi \}.
\]
We call $C$ a set of hard primitives when we enforce this restriction regardless of the evidence.

\medskip
\textbf{Setup.}
Let $\mu$ be a viability statement encoding the most basic valence partition.
Viable versus not viable.
For any policy $\pi$, define the viable continuation set
\[
\Omega_\pi := E_\pi \cap E_\mu.
\]
This is the set of viable future completions still compatible with $\pi$.
Define the free energy proxy in bits
\[
F_2(\pi) := \log_2 |E_\mu| - \log_2 |\Omega_\pi|.
\]
This equals $D_{\mathrm{KL},2}(q_\pi\|p_\mu)$.
Here $q_\pi$ is uniform on $\Omega_\pi$ and $p_\mu$ is uniform on $E_\mu$.

\medskip
\textbf{Theorem 3 (The Psychophysical Principle of Causality).}
Assume $C\neq\emptyset$ contains at least one commitment that is not forced by viability.
This means there exists $o\in E_\mu$ such that $c\not\subseteq o$ for some $c\in C$.
Then any policy that is forced to include $C$ has strictly higher $F_2$ whenever $C$ excludes at least one viable continuation that would otherwise remain possible.
So a generalisation optimal learner cannot keep such non valence primitives fixed.

\medskip
\textbf{Proof sketch.}
Adding a commitment intersects the viable continuation set with a smaller extension.
That can only shrink $\Omega_\pi$.
Since $F_2(\pi)$ is monotone in $|\Omega_\pi|$, the free energy increases strictly if any viable continuation is removed.
See ESM Theorem 3 for the full proof.

\medskip
\textbf{Interpretation.}
The only safe primitive is the qualitative viability signal.
That is what we call valence.
Quality neutral representations must therefore be constructed later, as learned causes of valence.
\end{widegraybox}

\section{Multi-Layered Self-Organisation}

As the vocabulary and capacity for w-maxing scale, a greater variety of concepts can be learned. Progressively higher orders of ‘causal-identity’ for one's self related information processing are constructed \cite{bennett2023c}. This lets us frame the construction of embodied selves in developmental \cite{ciaunica2019} and evolutionary terms. 

\subsection{The First Order Self}

A first order self (1ST henceforth) allows an organism to discern the consequences of its actions.
In our formalism it is the weakest causal identity that distinguishes self generated interventions from matched observations.
The full definition and proof are in the ESM.
This serves as the locus of self related information processing and experience \cite{bennett2023c}, allowing the organism to plan complex interactions and maintain a consistent ``self'' that is part of the present interaction. 1ST order selves amount to reafference \cite{merker2007,barron2016}, observed in mammals and insects.

\paragraph{A formal role for the 1ST order self.}
If an organism can improve expected utility by treating intervention episodes differently from observation episodes, then some internal variable must encode that difference.
Otherwise the organism would be forced to map the same embodied statement to the same internal state, and so could not act differently in the two cases.

Under the scale and incentive preconditions already described, the candidate set of self intervention signatures is nonempty.
A w-max learning rule then selects a maximally weak signature from that set.
Any w-maximiser is inclusion minimal among candidates.
If a strictly smaller signature existed, it would have a strictly larger extension and would have been preferred.
This is the operational content of reafference in our framework.

\paragraph{Why self is a fixed point of w-maxing.}

A short proposition shows that if there exists a minimal self intervention tag that is implied by every viable policy, and that tag is itself viable, then it is the weakest viable policy and w-maxing converges to it.
See Box 3 for the sketch and see the ESM proposition titled ``Self fixed point'' for the full proof.

A 1ST order self may require centralisation and a ``solid brain'' with a persistent structure \cite{sole2019} (an individual human), as opposed to a ``liquid brain'' (e.g. an ant colony or a population of humans). A solid brain would be required for the signals that contribute to a first order self to be co-instantiated in proper time \cite{bennett2026a}, and causally affect one another \cite{bennett2026c}.

\subsection{The Second Order Selves}

Survival may demand organism $\mathfrak{a}$ infer $\mathfrak{b}$'s prediction of $\mathfrak{a}$'s interventions (to see one's self as if through another's eyes \cite{bennett2023d}). This is called a second order self (2ND henceforth). 
We argue that if access conscious contents are available for \textbf{communication} in the human sense, then they must be communicable in the Gricean sense \cite{grice1957,grice1969}. Grice argued that communication is about the inference of intent. If person $\mathfrak{a}$ and $\mathfrak{b}$ are talking, then the meaning $m_\mathfrak{a}$ of what $\mathfrak{a}$ says is whatever $\mathfrak{a}$ intends $\mathfrak{b}$ understand. The meaning $m_\mathfrak{b}$ that $\mathfrak{b}$ understands is whatever $\mathfrak{b}$ thinks $\mathfrak{a}$ wants $\mathfrak{b}$ to think. $\mathfrak{b}$ has understood what $\mathfrak{a}$ means if $m_\mathfrak{b}$ approximates $m_\mathfrak{a}$. This can happen only if $\mathfrak{a}$ can predict with reasonable accuracy what $\mathfrak{b}$ thinks $\mathfrak{a}$ thinks, and $\mathfrak{b}$ can predict what $\mathfrak{a}$ thinks $\mathfrak{b}$ will think upon hearing an utterance. In other words, both $\mathfrak{a}$ and $\mathfrak{b}$ must have 2ND order selves that are good approximations. Yes, there are other aspects to communication. 

However, here we are talking about consciousness. Access conscious contents are those available for reasoning and report. It follows\footnote{If the definition of access consciousness is to be consistent with reasoning and report as exhibited by conscious humans.} that access conscious contents must in principle be communicable in the sense Grice described.

As such, we argue contents available for communication can only be the contents of 2ND order selves, which means only an organism with 2ND order selves can be considered to have access consciousness. 2ND order selves also explain attention and self-awareness. An organism can have many 2ND order selves because they depend upon who or what the organism is interacting with, just as the availability of information depends on context. 

Where a 1ST order self might allow one to observe a cat and form plans regarding causal interactions with the cat, a 2ND order self would allow one to be consciously \textit{aware} of the cat for the purpose of reasoning and report. One can know of the cat, and that another organism knows of the cat, but a 2ND order self is insufficient to be aware that one is aware of the cat\footnote{This is `meta-self-reflexive consciousness' as some have described it \cite{morin2006}.}.

More formally using the notation given in the appendix \cite{bennett2025thesis} (see quick reference guide in section 2), assume $\mathfrak{a}$ and $\mathfrak{b}$ are organisms that evolved to accurately predict one another's behaviour.
Assume $\mathfrak{a}$ constructs a causal identity $c^\mathfrak{b}_\mathfrak{a}$ to predict $\mathfrak{b}$ given input $i_\mathfrak{a} \in I_{\mu_\mathfrak{a}}$, of which a second order self $c^\mathfrak{ba}_\mathfrak{a}$ is part. Likewise, $\mathfrak{b}$ constructs $c^\mathfrak{a}_\mathfrak{b}$ to predict $\mathfrak{a}$ given input $i_\mathfrak{b} \in I_{\mu_\mathfrak{b}}$, of which $c^\mathfrak{ab}_\mathfrak{b}$ is part.
What is important here is that each organism's intent is to some extent inferred by the other, changing the sorts of policies that are fit.
For example, second order self means each knows the other can anticipate manipulation, which means the optimal policy will often be to \textit{have} rather than feign intent that aligns to some extent with the other party's desires, to co-operate \cite{alexander2022}\footnote{Depending upon circumstances, for example organisms may co-operate in some circumstances but not others \cite{bennett2023d}, and transient relations, information asymmetry and other factors can make deceit a more attractive option.}. Repeated interaction creates an iterated prisoner's dilemma, incentivising co-operation and signals that both parties interpret similarly (the beginnings of language) \cite{bennett2023d}. 
To communicate in Gricean terms, $\mathfrak{a}$ must intend to convey meaning $m_\mathfrak{a}$, and $\mathfrak{b}$ must recognise this intent. The incentive precondition explains \textit{why} $\mathfrak{a}$ would form such intent (co-operation is often advantageous), while \textit{how} may be understood as follows: \begin{itemize}
    \item $c^\mathfrak{ba}_\mathfrak{a}$ lets $\mathfrak{a}$ predict what $\mathfrak{b}$ will come to believe when it observes $\mathfrak{a}$'s behaviour. 
    \item $c^\mathfrak{ab}_\mathfrak{b}$ then lets $\mathfrak{b}$ predict what $\mathfrak{a}$ intends that $\mathfrak{b}$ believe.
\end{itemize} 
$\mathfrak{a}$ can use $c^\mathfrak{ba}_\mathfrak{a}$ to infer behaviour to which $\mathfrak{b}$ will ascribe the intent to communicate $m_\mathfrak{a}$, and $c^\mathfrak{ab}_\mathfrak{b}$ lets $\mathfrak{b}$ infer that this is what $\mathfrak{a}$ intends. The ``utterance'' Grice refers to is how $\mathfrak{a}$ affects $\mathfrak{b}$ in accord with earlier definitions.
Put another way, $\mathfrak{a}$ \textit{encodes} $m_\mathfrak{a}$ into its behaviour in a manner that $\mathfrak{b}$ can \textit{decode} (their respective second order selves act as encoders and decoders). By encode and decode, we mean a loose approximation of $m_\mathfrak{a}$ is communicated. There are of course shortcuts, for example of $\mathfrak{a}$ and $\mathfrak{b}$ are of the same species then they likely have similar motives and experiences, and so the efficient thing for each to do would be to use its own intent as an approximation of what the other might think. However, that does not obviate the need for second order selves, it just makes them easier to realise. Formal and experimental results supporting our claims regarding 2ND order selves are given in the ESM.

\paragraph{A minimal separation task.}
Consider a speaker who must transmit a private bit to a listener.
The listener can be in one of two decoder modes.
One decodes literally and the other flips the bit.
If the speaker cannot condition on any evidence about which decoder is active, its success is at most one half.
If the speaker can send one probe signal and observe the listener's response, it can infer the decoder and succeed with probability one.
That conditioning variable is the smallest possible second order self.
It is an internal handle for how the audience is interpreting the speaker.

\paragraph{Decoder mismatch in Monte Carlo benchmarks.}

Monte Carlo benchmarks in ESM Experiment S2 show that a couple of short probes largely eliminate the cost of decoder mismatch.
Passive signalling stays near chance, while probing approaches oracle performance.
See Box 4 for the headline results, and see ESM Experiment S2 for the full protocol and plots.

\subsection{The Third Order Selves}

We can scale preconditions indefinitely.
For the purposes of this paper we stop at 3RD order because this is where a self can extend across time as a stable narrative.
It is where an agent can treat its present actions as evidence about its future behaviour.
It is also where the agent can deliberately bind itself to reduce what it could do later.
This is what makes long horizon cooperation and trust possible.

Formally, a 3RD order self for $\mathfrak{a}$ relative to an audience $\mathfrak{b}$ is the iterated causal identity
$$
c^{\mathfrak{baba}}_{\mathfrak{a}}.
$$
It is $\mathfrak{a}$'s prediction of $\mathfrak{b}$'s prediction of $\mathfrak{a}$'s prediction of $\mathfrak{b}$'s prediction of $\mathfrak{a}$.
$\mathfrak{a}$'s action now changes what $\mathfrak{b}$ expects of $\mathfrak{a}$ later, and $\mathfrak{a}$ knows that $\mathfrak{b}$ knows that $\mathfrak{a}$ knows.

A formal separation in the ESM shows why narrative, 3RD order selves matter for trust.
Even with perfect audience modelling, nonbinding promises do not support trust in equilibrium when exploitation remains available.
If an agent can publicly bind itself by paying a one time cost that removes an exploit option, a trust equilibrium exists.
Monte Carlo benchmarks in ESM Experiment S3 show the same pattern across randomised trust games.
See Box 3 for the formal sketch and Box 4 for the experimental headline results.
See ESM Experiment S3 for the full protocol and plots.

\begin{widegraybox}[Other formal proofs in support of the main text claims][label=box:proofs]
The main text uses proof results besides Theorem 3. They constrain and support the narrative arguments of this paper.
Here we give only the sketch level version. Complete proofs are in the ESM.

\medskip
\textbf{Reminder.}
Weakness maximisation, called w-maxing, means selecting a correct policy with maximal weakness $w(\pi)=|E_\pi|$.
In plain English, it keeps the largest set of future refinements open while staying correct. To tie this to related biological frameworks on cognition spaces \cite{sole2026cognitionspaces}, weakness corresponds to the size of the ``cognitive light cone'' of an embodied system.

\medskip
\textbf{Result 1. w-maxing learns causal intervention tags.}
Given matched intervention and observation episodes, a w-maximised causal identity exists in any finite embodied language and is inclusion minimal among candidates.
In plain English, it is the least committal signature that still separates self caused change from external change. This proves an earlier but less formal claim \cite{bennett2023c}.

\medskip
\textbf{Result 2. Decoder mismatch can hide minds.}
A Bayesian likelihood ratio bound shows that an evaluator can be forced toward a noise hypothesis when every mind-like hypothesis in its model class fits the observed trace worse than noise by a constant factor.
In other words, if you decode the signal with the wrong key, Bayes rule will tell you there is no sender\footnote{This creates mind-blindness and relates to The Fermi Paradox \cite{bennett2022b}.}. 

\medskip
\textbf{Result 3. The 1ST order self is a w-max fixed point.}
If there exists a minimal self intervention tag that is implied by every viable policy and is itself viable, then it is the weakest viable policy and a w-max learner converges to it.
In plain English, reafference becomes a stable self tag when survival depends on it. It is a mathematical result that \textit{predicts} the reafference we observe in living organisms.

\medskip
\textbf{Result 4. Gricean report and trust force higher order selves.}
In Gricean communication, what counts as a correct report depends on the listener's inference, not a fixed signal to world mapping.
We show that any signalling policy that is robust across variation in an audience's epistemic state must condition on an audience model.
In our formalism, that audience model is a 2ND order self.
We also give a minimal commitment game in which even perfect audience modelling cannot sustain trust when exploitation remains available.
If an agent can publicly bind itself by paying a one time cost that removes an exploit option, a trust equilibrium exists.
In other words, reportable access is not just the ability to say a thing.
It is saying the thing knowing a listener will understand, by predicting what they predict you intend by what you say, and sometimes even binding yourself so they can believe you (because they will predict your self bind and observe your behaviour is consistent with it).

\medskip
Taken together, \emph{Results 3 and 4 make the self hierarchy a derived necessity}.
It is the minimal structure that survives under generalisation optimal adaptation when the niche includes self intervention, report, and trust. See the ESM propositions on causal identity minimality, decoder mismatch, self fixed points, Gricean report, and credible commitment.
\end{widegraybox}

\section{The What and Why of Consciousness}

Up to now we have developed the conceptual toolbox that we use to cross the ``explanatory gap'' and explain consciousness from an embodied perspective. We started with the observation that a human body is a system that self-organises to maintain itself in the face of change both within and without its finite boundaries. To achieve this vital goal, the embodied system needs to process information qualitatively, meaning it is intrinsically valanced in service of survival. An organism does not spring into existence ``objectively'' by understanding objects and properties, but rather constructs them through subject-oriented interaction with the environment. One can regard objects and properties as policies according to which the environment is subjectively interpreted. To learn how to interpret the environment an organism must differentiate between states, must react to change, and learn policies according to the valence associated with that change. But how do we get from these embodied subjective experiences to the subjectivity that is usually linked to phenomenal consciousness in awake adult minds?

We will now argue the following.
\begin{enumerate}
    \item There is something it is like to be an organism with a 1ST order self.
    \item Subjective experience is the process of learning and enacting a hierarchy of causal identities.
    \item Access consciousness requires both 1ST and 2ND order selves.
\end{enumerate}

Because we do not presuppose representations, we explain how qualitative \textit{interpretations} emerge from attraction to and repulsion from contentless physical states. This means the foundation of information processing in biological self-organising systems is valence.
For biological, living self-organising systems, being alive is intrinsically good, being dead or ill is bad. For example, in developing self-organising systems such as the human body, it is good to have a better developed sense of smell than vision at early stages of life (babies are more accurate smellers than adults \cite{schaal2020}). Different senses have different degrees of valences at different stages of the developmental ladder. Hence there is ``something what it is like'' to be a baby which is different from something what it is like to be an adult. However, the idea here is that basic experiences (i.e. the bodily information processing subserving the ultimate goal of staying alive) have valence at their very core (good=stay alive; bad=dead). This culminates in experiential, phenomenal `quality' of appearance, of what it is likeness and how thing appear to the system (i.e. phenomena). That is the aspect that philosophers typically tackle. Our aim is not to dismiss the latter; our claim is that one cannot have the latter without the former.

As we saw earlier, learned policy has valence, and it is a classifier of inputs. Those inputs are what one may call ``information'' in the mechanistic sense of the environment being in one state and not another. Hence, a policy is a classifier of information, but that information is not in a language, and it is not yet something labelled or quantified (until there is a policy that labels or quantifies).

Clearly, prelinguistic self-organising systems must classify and attach value and disvalue to states and anticipated states to prioritise and make decisions. To learn is to sense ``something'' and, motivated by valence, construct a policy classifying that ``something''. When we say physical states are contentless, we are not suggesting physical objects and properties don’t exist. Objectively, a physical object is an aspect of the environment and every aspect is realized by one state or another. However, previous work has shown that for every aspect of the environment there is an equivalent program, meaning from an objective point of view what is or is not a physical object or property is a matter of interpretation \cite{bennett2024b}. The radical and novel claim here is that, subjectively, an object or property only exists to the extent that that organism constructs a causal identity for it. 

At the most basic level, the foundation of all of this is physical attraction and repulsion. An organism hard-wired by natural selection can be physically attracted or repelled without any concept of what it is physically attracted or repelled from. The system doesn't know to what it is attracted or repelled from in the sense that it has not constructed a causal identity for that object. There is no named object which has the property `attractive’. The organism itself is simply attracted or repelled according to hard-wired policy. For example, a single-celled organism might be attracted to glucose, and its response to `tumble’ or `stop’ would be hard-wired. It does not satisfy the incentive or scale preconditions to construct a causal identity for glucose, or `sweet’, or hunger, or anything else. Intuitively, one might think of this like `one dimensional’ valence. To the bacteria, there is no subjective categorical variable or object to which it is attracted, though in the objective sense such a thing does exist. It is simply attracted or repelled.

Over the course of this paper we have sought to explain how we get from simple `one dimensional’ hard-wired valence to a property or quality that has valence.  We suggest that categorical variables for properties or `qualities’ like `the colour red’ or `the smell of coffee’ only exist when we have a causal identity for them, which requires scale and incentive. It can be hard-wired or learned. When we have only one cell we do not have a vocabulary capable of satisfying the scale precondition. However self-organising biological systems are collective \cite{fields2020}. Control is delegated and distributed among smaller components \cite{bennett2024c} which operate concurrently. An individual cell might only have `one dimensional’ valence\footnote{Debatably, as even single cells have complex sub-components that interact with each other.}, but as soon as we have two cells we have something analogous to a second `dimension’. Cellular collectives are polycomputational systems, meaning one cell can play a part in more than one computation simultaneously, at different scales \cite{bongard2023}. 

As we scale up the collective, it is impelled by the cells of which it is formed. Instead of `one dimensional’ valence, we now have a rich tapestry of competing drives \cite{bennett2025thesis}\footnote{The contributing parts of this tapestry are co-instantiated in \textit{proper} time \cite{bennett2026a} and causally influence one another in space \cite{bennett2026c}. This is important to understand how it relates to notions of conscious ``unity'' and ``integration'' in other theories \cite{dennett1992,varela2001,singer1995,fries2005,baars1988,dehaene2001,tononi2004,lamme2006} and psychophysics \cite{eagleman2000,poppel1997,vroomen2010,wallace2014}.}. Our position is that it is this rich tapestry that is a property or `quality’ of a state of the environment interpreted by the organism. It is this rich tapestry that can be explained through causal identities for properties like `the color red’, `hunger’ or `thirst’. A higher level of abstraction may be formed from this rich tapestry of valence and how this changes over many interactions (the different inputs and outputs in a task that imply generalizable causal identities). For example, hunger and thirst might have the same overall intensity and so one could say they have the same overall valence, but they are qualitatively different because they are different `tapestries’ of valence at a lower level of abstraction. This is what is meant by The Psychophysical Principle of Causality in Section 3. 

In a self-organising system of sufficiently complex scale with sufficient incentive\footnote{Here we refer to the scale and incentive preconditions formally defined earlier.}, what starts out as simple `one dimensional’ valence culminates in categorical variables that serve different organismic needs. This is why the sound of middle C is phenomenologically different from the color blue. They have different causal identities. Middle C results from sound waves hitting the ear, blue results from light waves hitting the eye. These variations produce different subjective embodied experiences. 

The bold claim here is that information processing (i.e. exploring one’s body and environment) in self-organising systems such as the human body is necessarily qualitative. Note that this is different from experience in the sense usually defined in the literature. 

To put it provocatively quality precedes quantity, and quantity is nothing more than the interpretation of quality. Quality comes first and experiences should be regarded on a continuum rather than a switch on / switch off phenomena. All living systems experience the world through their bodies and as such, there is something it is like to experience the world in that basic way (even when one is asleep). One can access those phenomenal, experiential aspects at a higher level, true, but by accessing them, it doesn’t mean that one `constructs’ consciousness or one becomes a conscious being. One is already consciously experiencing the world before one can explicitly access one’s own experiences in 2ND order selves. 

It follows that every policy learned in this way must classify a quality. Hence every such policy is a local state. A causal identity is inherently dispositional. It is a product of valence in self-organising systems such as human bodies. There would be a policy for one's act of smelling coffee. For perceiving one's friend. There would be something different it is like to interact with a hostile version of that very same friend. The 1ST order self accompanies everything an organism does. It has a quality, so the 1ST order self is ``\textbf{what it is like}'' to be that organism. Put another way, Nagel's question of ``what it is like to be'' a particular organism could be answered if one could somehow have that organism's first order self \cite{nagel1974}. An organism does not make a decision to interpret information, it just does it.

What we call \textbf{subjective experience} is the process of learning and enacting a lattice of causal identities, and there is something that ``has'' all of those subjective experiences once there is a 1ST order self that is part of them all. This is where \textbf{phenomenal consciousness} begins, with subjective embodied experience as learning and exploration.

Likewise one's 2ND order selves would have a certain qualitative character, and one's 3RD order too. 

There would, however, be a very clear delineation between the presence and absence of conscious subjective experience. The absence of a 1ST order self would mean there would be no policy linking all interventions together, and so no ``self'' that experiences making them. Hence a 1ST order self must precede higher order selves and not the other way around. Our approach thus stays in stark contrast with Higher Order Thought Theories of Consciousness (HOT) \cite{fleming2020awareness}.

Rather our arguments align with those of Merker, who has linked subjective experience to reafference \cite{merker2005,merker2007,barron2016}. We agree that reafference is key, but we provide a very different explanation of why and how. Their work presents biological evidence for subjective experience in organisms with reafference. In contrast, we derive the 1ST order self from first principles and explain why and how it is a classifies ``what it is like'' to be a particular organism. The 1ST order self also happens to be equivalent to reafference, so we arrive at the same conclusion as Merker from very different, mathematical premises. Hence these are complementary positions.

\section{Collapsing The Abstraction Layers}

Our theory of embodied consciousness places the subjective experience of embodied organisms in the physical world and not in some parallel, dualist world. This means roughly that the psychophysical principle of causality takes us from attraction to and repulsion from physical states, to higher order meta representations. This produces different orders of self. We suggest that the full richness of human subjective experience depends upon these different orders of self\footnote{Perhaps the full richness of ``hard'' consciousness \cite{boltuc2012} depends upon the interaction of selves, perhaps in planning \cite{shanahan2012}.}.
Conscious ``access'' as a human has it involves \textit{meaningful} report, which as we've established requires 2ND order selves. This is very different from the mere ``access'' to information, in the sense of impinging signals upon the system or stored in memory. 
However, supposing we could somehow contrive a system with 2ND order selves but no 1ST, then 2ND order selves alone would still not allow one to reason about how one's interventions might affect the contents of 2ND order selves to communicate meaning. For that, one requires a 1ST order self. Hence access consciousness requires both 1ST and 2ND order selves. 

Our radical and provocative claim is that phenomenal consciousness without access consciousness is likely very common, but the reverse is implausible. Our framework thus makes a philosophical zombie impossible because there can be no perfect unconscious replica of a conscious embodied organism but without the qualitative aspect of information processing attached to it\footnote{Removing the qualitative aspects would introduce unnecessary extra steps and thus inefficiency\cite{bennett2025thesis}.}. 
In a self-organising system is causality all the way down. Layer upon layer of abstraction classifies the causes of valence across spatial and temporal scales, from simple one-dimensional signals up to complex tapestries of valence that are the phenomena that make up a self-organising system's experience.

Related work argues qualia are nothing more than the tapestries of valence we describe \cite{bennett2025thesis}. Put another way, it claims qualia are reducible to valence. If one were to subscribe to this view, then one might conclude it dissolves hard problem of consciousness, because the formalism\footnote{Stack Theory, and Pancomputational Enactivism within it.} describes every conceivable world and implies a philosophical zombie with access but not phenomenal consciousness is impossible in all of them. A philosophical zombie acts exactly as a human but without qualia. If one takes qualia to be tapestries of valence, then a system with a 1ST-order-self and the efficient, qualitative information processing we have described would have qualia. Thus a human without qualia would be less efficient than its conscious counterpart and thus not strictly speaking a philosophical zombie.

\section{From Rocks to Einstein: The Hierarchy of Being}
In the last section of the paper we suggest the problem of phenomenal consciousness has it backwards. The question is not why qualia exist, but why anyone thinks representational contents can exist in a prelinguistic organism without direct modelling of qualities (i.e. qualia) and discrimination via information processing of cause and affect.
To illustrate how our argument applies in the real world we need to go back to the simplest basic level and describe stages of conscious organism. Each stage follows from scaling up supply and demand for w-maxing, through natural selection. For each stage we point out animals which are likely to be \textit{at least} so conscious:\\

$0:$ Inert \textit{(e.g. a rock)}

$1:$ Hard Coded \textit{(e.g. protozoan)}

$2:$ Learning \textit{(e.g. nematode)}

$3:$ 1ST Order Self \textit{(e.g. housefly)}

$4:$ 2ND Order Selves \textit{(e.g. cat)}

$5:$ 3RD Order Selves \textit{(e.g. human)}

\paragraph{Stage 1: Hard coded.}  
Stage one refers to adaptations hard-wired by natural selection, allowing complexity to persist \cite{heylighen2023b} in a stable environment. 

\begin{itemize}
    \item \textit{What:} Hard-coded adaptations. Habituation and sensitization.
    
    \item \textit{How:} The extension of fit behaviour is learned by natural selection and hard-coded into the organism as a policy (in DNA, form, the local environment etc).
    
    \item \textit{Why:} If the environment is very predictable, it may be more efficient to hard-code fit behavior.

    \item \textit{Example:} Single-celled protozoan. 

\end{itemize}

\paragraph{Stage 2: Learning.}  
Stage two introduces learning. To learn an organism must store, classify and order historical examples by valence. However there is not ``something it is like'' to be stage two, because there is no locus of ``self''. A biological example of such a decentralised nervous system is the cubozoan box jellyfish Tripedalia cystophora. Even Tripedalia cystophora was recently shown to be capable of associative learning \cite{bielecki2023}. An entirely distributed control system can ``learn''. Likewise, stage two is exemplified by nematodes \cite{rankin2022,willett2018} with a centralised nervous system and some ability to adapt with experience. However, the absence of a ``self'' limits causal reasoning, which as others have already pointed out must limit spatial, navigational abilities \cite{barron2016}. When starved C. elegans exhibit ``increased locomotion and dispersal in a random, rather than directed, search'' \cite{barron2016,luersen2014,artyukhin2015}, whereas something like a bee or an ant can recall and navigate to previously discovered food  \cite{wehner2013,seeley1995page295,oades1978}.

\begin{itemize} 
    \item \textit{What:} Learning.
    
    \item \textit{How:} Valence. 
    
    \item \textit{Why:} An organism that can learn can survive in more circumstances than one that cannot.

    \item \textit{Examples:} Jellyfish, nematode. 
\end{itemize}

\paragraph{Stage 3: 1ST order self.}
This is where phenomenal consciousness begins, with a 1ST order self. In biological terms this implies reafference, which others have argued is the key to subjective experience, albeit for different reasons than what we have \cite{merker2005,merker2007,barron2016}. They identified a housefly as a good example of where subjective experience may begin, and we concur.
We also hold this is where an organism might be said to have intent. Intuitively, the policy that motivated behaviour is the intent of that behaviour\footnote{In the same way declarative and imperative programs are equivalent \cite{howard1980}.}. For example, eating tends to involve the intent of satisfying hunger, which in turn satisfy the basic policy `stay alive'.
A stage three organism can feel simple things like hunger, but cannot conceive of itself from another's perspective. Subsequently it cannot communicate in the Gricean sense \cite{grice1969,bennett2023d}, or conceive of its own death, or experience shame.

\begin{itemize}
    \item \textit{What:} 1ST order self. Reafference. Phenomenal consciousness. 
    
    \item \textit{How:} Embodiment in which intervention is not identical to observation. 
    
    \item \textit{Why:} Accurate prediction of consequences of interventions. For example, a fly must distinguish between having moved, and the environment having moved, to navigate. 

    \item \textit{Example:} Housefly. 
\end{itemize}

\paragraph{Stage 4: 2ND order selves.}
Stage four is the 2ND order self, and this is where Block's reportable access consciousness begins in our formal account, because information is now available for report in the Gricean sense. In the ESM we prove that robust Gricean report requires an audience model, which is exactly a 2ND order self. The ability of ravens to intentionally deceive \cite{bugnyar2002} suggests they are \textit{at least} stage four. Raven $\mathfrak{a}$, aware that it is being observed by raven $\mathfrak{b}$, will act as if it is hiding food in one location to mislead $\mathfrak{b}$, but will then move the food in another location unobserved by $\mathfrak{b}$. $\mathfrak{a}$ seems to predict not just the intent of $\mathfrak{b}$ (to steal the food), but $\mathfrak{b}$'s perception of $\mathfrak{a}$. It seems likely that dogs and cats have second order selves, as they must hunt reasonably intelligent animals and must anticipate how their actions are perceived. For example, a cat anticipates its prey will flee when it is observed, and hides.

\begin{itemize} 
    
    \item \textit{What:} Reportable access consciousness. Theory of mind. Self-awareness. 
    
    \item \textit{How:} Selection pressures that demand theory of mind. 
    
    \item \textit{Why:} A 2ND order self is necessary to infer, communicate and manipulate intent. 

    \item \textit{Example:} Cats, dogs, ravens.
\end{itemize}

\paragraph{Stage 5: 3RD order selves (narrative selves).}
A 3RD order self is a narrative self.
It lets an organism treat itself as persisting across time, with a story that links past interventions to future commitments.
Formally, a 3RD order self is a 2ND order self for one's 2ND order self.
In the ESM we give a minimal trust and commitment game that separates 2ND from 3RD order selves.
It shows that even perfect audience modelling cannot sustain trust unless a publicly legible self binding move exists.
Humans appear to possess this.
Altruistic behaviour observed in Australian magpies \cite{potvin2022} suggests they may also have narrative selves.

\begin{itemize}
    \item \textit{What:} Impelling narrative self. Meta self-awareness. Trust. 
    
    \item \textit{How:} More accurate prediction and planning.

    \item \textit{Why:} Because a social organism must predict complex social dynamics. 

    \item \textit{Example:} Human. Highly intelligent animals such as Australian magpies may also be this conscious. 
\end{itemize}

\begin{figure}
\centering
\includegraphics[width=.8\linewidth]{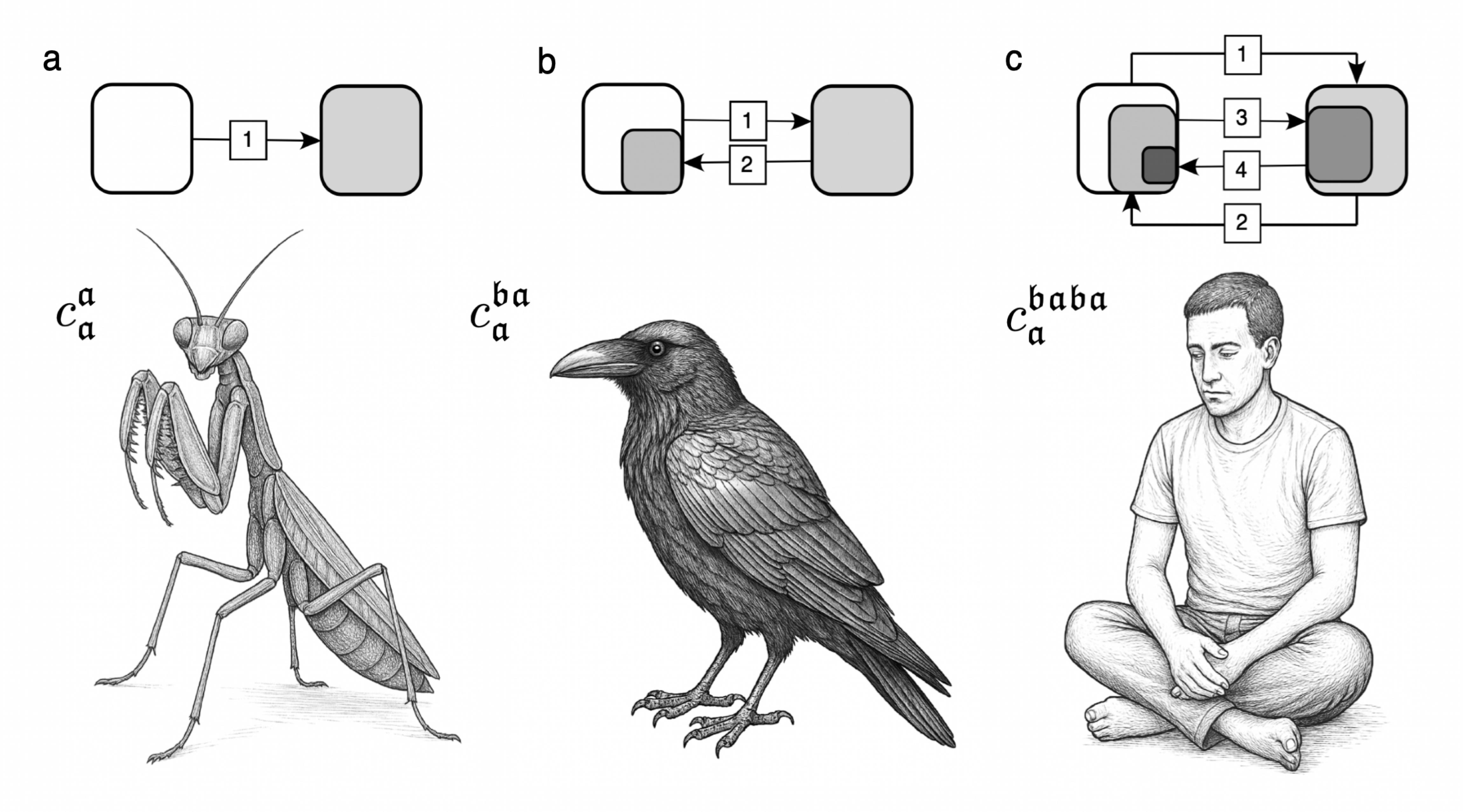}
\begin{tikzpicture}[node distance=0.32cm]
\tikzset{
  selfbox/.style={draw, rounded corners=6pt, fill=gray!5, align=left, text width=0.9\linewidth, inner sep=6pt, font=\small},
  selftitle/.style={font=\bfseries\large, align=center, text width=0.9\linewidth},
  selfnote/.style={font=\footnotesize, align=left, text width=0.9\linewidth}
}
\node[selftitle] (t) {Hierarchy of Selves};
\node[selfbox, below=0.25cm of t] (b0) {\textbf{0TH. No self} \\ No stable self tag. Behaviour is reactive and not model based.};
\node[selfbox, below=0.25cm of b0] (b1) {\textbf{1ST. Self tag} \\ Separates self generated interventions from matched observations. This is reafference. Grounds phenomenal consciousness.};
\node[selfbox, below=0.25cm of b1] (b2) {\textbf{2ND. Audience model} \\ Models how an audience will decode the agent. Enables Gricean report. Grounds reportable access consciousness.};
\node[selfbox, below=0.25cm of b2] (b3) {\textbf{3RD. Narrative self} \\ Stabilises an identity across time. Enables long horizon planning, self binding, trust, and an impelling narrative.};
\node[selfbox, below=0.25cm of b3] (n) {\textbf{NTH. Order selves} \\ Higher order selves may facilitate as yet unexplored capabilities.};
\end{tikzpicture}
\caption{Overview of orders of self as nested causal identities. 0TH is an embodied agent without an explicit self tag. 1ST adds a reafference like separator between self generated interventions and matched observations. On our account this grounds minimal phenomenal consciousness. 2ND adds a model of how an audience will decode the agent. This enables Gricean report and grounds reportable access consciousness. 3RD is a narrative self that stabilises an identity across time. This enables long horizon planning, self binding, trust, and an impelling narrative. Formal definitions and proofs are in the ESM, where we derive these orders from generalisation optimal learning under the scale and incentive preconditions. Picture kindly provided by Ricard Solé of The Santa Fe Institute.}\label{fig1}
\end{figure}

\section{Conclusions: Why Nature Does Not Like Zombies}
In this paper we have proposed a novel, causal identity theory of consciousness, taking as a starting point not the conscious mind but the embodied experiencing organism. Building on Stack Theory \cite{bennett2025thesis}, we have illustrated in formal terms how biological self-organising systems become phenomenally conscious when they construct a 1ST order self. A human lacking a 1ST order self could not perform causal reasoning needed to adapt as healthy humans can \cite{pearl2018}.

\begin{widegraybox}[Experimental results at a glance][label=box:experiments]
We report three simple Monte Carlo benchmarks from the associated workshop paper \cite{bennett2026b}.
They are minimal tests of the formal claims in noisy randomised settings.

\medskip
\textbf{Experiment 1. Weakness beats simplicity.}
Across randomised abstraction layers, w-maxing has zero regret under a natural ignorance prior.
A shortest description length selector incurs large regret.
In one benchmark, that regret is about $13.4$ bits.
That is an odds penalty of about $2^{13.4}\approx 10{,}800$.
In other words description length depends on the encoding, while weakness is an invariant of the embodied language.

\medskip
\textbf{Experiment 2. Probing defeats decoder mismatch.}
When a listener's decoding rule is hidden, passive signalling stays near chance.
With two short probes, a probing policy reaches about $0.93$ success across both stationary and nonstationary decoder settings.
An oracle decoder reaches about $0.96$ in the stationary setting.
In other words, causal intervention helps ascertain the listener's decoder for communication.

\medskip
\textbf{Experiment 3. Binding creates trust.}
In randomised trust games, nonbinding promises do not generate trust in equilibrium.
Binding moves do.
In one benchmark sample, binding yields trust about $0.95$ by horizon $H=5$. Third order selves thus help navigate complex social environments.

\medskip
\textbf{Protocols and plots.}
See ESM Experiments S1 to S3, with Tables S1 to S4 and Figures S1 to S4.
\end{widegraybox}
One consequence of our approach is that it places the qualitative aspect of conscious embodied experience\footnote{Inherently valenced information processing through the lens of self-preservation.} before access consciousness. Ontologically, this means that before one can access information, one needs to experience information through bodily and causal interaction with the world. We have shown a consistent definition of access consciousness requires 1ST and 2ND order selves. Phenomenal consciousness arises first, access comes later. Unlike panpsychism, we don't believe rocks are conscious. Only self-organising systems that need to adapt, motivated by valence, while keeping track of self-relevant information are conscious. Consciousness is a key adaptation of a vulnerable organism in a physical world. 

In summary, we have argued that access consciousness at the human level is impossible without the ability to hierarchically model the following.
\begin{enumerate}
    \item The self.
    \item The world and others.
    \item The self as modelled by others.
\end{enumerate}
These follow from learning and enacting a lattice of causal identities, if the scale and incentive preconditions are met. There is something it is like to be an organism with a self. Paradoxically, phenomenal consciousness begins in organisms that have an end and a primary goal to avoid that end. Our proposal lays the foundation of a formal science of consciousness, deeply connected with natural selection rather than abstract thinking, closer to the facts of embodiment than the fiction of zombies.


\hypertarget{acknowledgements}{%
\section*{Acknowledgements}}

{\label{687807}}

Supported by a Fundaçao para a Ciencia e a Tecnologia (FCT) grant PTDC/FER-FIL/4802/2020; and 2020-02773 CEECIND FCT to A.C.

\selectlanguage{english}
\FloatBarrier

\bibliographystyle{plainnat}
\bibliography{master_bibliography}

\end{CJK}\end{document}


\maketitle

\section{What is in this ESM}
This document is the companion Electronic Supplementary Material for the main manuscript of ``Why Is Anything Conscious?''.
It does three things. First, it gives the precise definitions that the main text uses.
Every mathematical object is defined in math and then paraphrased in plain English. Second it contains full proofs, whereas in the main text the proofs are just sketches. Third, it contains full experimental protocol details and the extra plots. The contents are in part a restatement of earlier proofs and experiments from \cite{bennett2025thesis,bennett2024c,bennett2026b}. The primarily novel formal contribution unique to this paper is the separation theorems, formally showing the hierarchy of selves follows from generalisation optimal learning (something claimed but not so formally backed in earlier work), and the formalisation and proof of The Psychophysical Principle of Causality. Please refer to the GitHub for an even more comprehensive list of related proofs, the experiment code and so forth \cite{bennett_appendix}.

\paragraph{Notation bridge to the main text.}
The main text writes the extension of a statement $x$ as $E_x$.
In this ESM, we write $\Ext{x}$.
These are the same object. The main text defines weakness as the size of an extension.
In this ESM we keep that definition.
Sometimes we also report $\log_2$ weakness.
That is just weakness measured in bits.

\section{Stack theory primitives used in this paper}
\subsection{Environments, programs, vocabularies, and statements}

\begin{definition}[Environment, programs, and vocabularies]
An \textbf{environment} is a nonempty set $\PhiEnv$ of mutually exclusive states.
A \textbf{program} is any set of states $p\subseteq \PhiEnv$.
Let $\Prog := 2^{\PhiEnv}$ denote the set of all programs.
A \textbf{vocabulary} is any set $\mathfrak{v}\subseteq \Prog$.
Unless we say otherwise, vocabularies are finite.
\end{definition}

\noindent\textbf{Interpretation.}
A state is a complete way the world could be. A program is a yes or no classification of states. It is a potential fact (we do not assume about what, it is merely a point of potential difference between states in the vein of structuralism). A vocabulary is the set of properties the body can stably express and use.

\begin{definition}[Abstraction layer, statements, and truth sets]
Fix a vocabulary $\mathfrak{v}$.
A \textbf{statement} is any subset $\ell\subseteq \mathfrak{v}$ whose truth set is nonempty.
The \textbf{truth set} of a subset $\ell\subseteq \mathfrak{v}$ is
\[
\Truth(\ell) := \bigcap_{p\in \ell} p,
\]
with the convention $\Truth(\emptyset)=\PhiEnv$.
The \textbf{language induced by $\mathfrak{v}$} is
\[
\Lang{\mathfrak{v}} := \bigl\{\,\ell\subseteq \mathfrak{v} \mid \Truth(\ell)\neq \emptyset\,\bigr\}.
\]
\end{definition}

\noindent\textbf{Interpretation.}
A statement is a bundle of properties that can all be true at once.
The truth set is the set of world states compatible with that bundle.
The induced language is the set of all bundles that are not contradictory.

\begin{definition}[Completions, extensions, and equivalence]
Fix a vocabulary $\mathfrak{v}$ and its language $\Lang{\mathfrak{v}}$. A \textbf{completion} of a statement $x\in \Lang{\mathfrak{v}}$ is any statement $y\in \Lang{\mathfrak{v}}$ such that $x\subseteq y$.
The set of all completions of $x$ is its \textbf{extension}
\[
\Ext{x} := \{\,y\in \Lang{\mathfrak{v}} \mid x\subseteq y\,\}.
\]
Two statements $x,y\in \Lang{\mathfrak{v}}$ are \textbf{extension equivalent} when $\Ext{x}=\Ext{y}$.
\end{definition}

\noindent\textbf{Interpretation.}
A completion is what you get by adding extra compatible details.
The extension is the set of all ways you could add details while staying consistent.

\begin{definition}[Weakness]
The \textbf{weakness} of a statement $x\in \Lang{\mathfrak{v}}$ is
\[
 w(x) := \card{\Ext{x}}.
\]
When we want a log scale we use
\[
 W(x) := \log_2 w(x).
\]
\end{definition}

\noindent\textbf{Interpretation.}
Weakness counts how many ways you can keep refining without breaking consistency.
Higher weakness means you have left more options open.
The log scale $W$ is just weakness measured in bits.

\subsection{Tasks, correct policies, inference, and learning}

\begin{definition}[$\mathfrak{v}$-task]
Fix a vocabulary $\mathfrak{v}$ and language $\Lang{\mathfrak{v}}$.
A \textbf{$\mathfrak{v}$-task} is an ordered pair $\alpha=\langle I_\alpha, O_\alpha\rangle$ where $I_\alpha\subseteq \Lang{\mathfrak{v}}$ is the input set and $O_\alpha\subseteq \bigcup_{i\in I_\alpha}\Ext{i}$ is the output set.
\end{definition}

\noindent\textbf{Interpretation.}
A task says which inputs can happen and which refined statements count as acceptable outputs.
It is not necessarily a function.
It is a constraint.

\begin{definition}[Inference and correct policies]
Fix a vocabulary $\mathfrak{v}$ and a task $\alpha=\langle I_\alpha,O_\alpha\rangle$.
A \textbf{policy} is any statement $\pi\in \Lang{\mathfrak{v}}$.
A policy $\pi$ is \textbf{correct} for $\alpha$ when
\[
\Bigl(\bigcup_{i\in I_\alpha}\Ext{i}\Bigr)\cap \Ext{\pi} = O_\alpha.
\]
Let $\Pi_\alpha$ denote the set of correct policies for $\alpha$.
The \textbf{inference problem} for $\alpha$ is to determine $\Pi_\alpha$.
\end{definition}

\noindent\textbf{Interpretation.}
A policy is a constraint that selects which outputs are allowed.
Correct means the allowed outputs match exactly the task outputs on the part of the world the task talks about.

\begin{definition}[Learning]
Let $\alpha$ and $\omega$ be tasks in the same language.
We say $\alpha$ is a \textbf{subtask} of $\omega$, written $\alpha\sqsubseteq \omega$, when $I_\alpha\subseteq I_\omega$ and $O_\alpha\subseteq O_\omega$. A learning system \textbf{generalises} from $\alpha$ to $\omega$ when it selects a policy $\pi\in \Pi_\alpha$ that is also correct for $\omega$, meaning $\pi\in \Pi_\omega$.
\end{definition}

\noindent\textbf{Interpretation.}
A parent task contains more cases than a child task.
Generalisation means a policy that fits the child also works on the parent.

\subsection{Causal identities and selves}

\begin{definition}[Causal identity candidates]
Fix an embodied language $\Lang{\mathfrak{v}}$.
Let $INT\subseteq \Lang{\mathfrak{v}}$ be a set of intervention episodes and $OBS\subseteq \Lang{\mathfrak{v}}$ a set of matched non intervention observations.
Define the candidate set
\[
\mathcal{C}(INT,OBS)
:=
\Bigl\{\,c\in \Lang{\mathfrak{v}}\setminus\{\emptyset\}\ \Bigm|\ 
(\forall\, int\in INT)\ c\subseteq int
\ \text{and}\ 
(\forall\, obs\in OBS)\ c\not\subseteq obs
\,\Bigr\}.
\]
Each $c\in\mathcal{C}(INT,OBS)$ is a causal identity candidate for whatever is common to the interventions but not already present in the observations.
\end{definition}

\noindent\textbf{Interpretation.}
A causal identity is a signature in the agent's own vocabulary.
It is present whenever the source intervenes.
It is absent in matched background observations.

\begin{definition}[w-maximised causal identity]
Assume $\mathcal{C}(INT,OBS)\neq\emptyset$.
A \textbf{w-maximised causal identity} is any candidate $c^\star\in\mathcal{C}(INT,OBS)$ that maximises weakness among candidates
\[
\card{\Ext{c^\star}} = \max\{\card{\Ext{c}} \mid c\in \mathcal{C}(INT,OBS)\}.
\]
When $\mathfrak{v}$ is finite, the maximum exists.
\end{definition}

\noindent\textbf{Interpretation.}
Many signatures might separate intervention from background.
W-maxing picks the one that leaves the most room for later refinement.

\begin{definition}[First order self]
Let an organism $\mathfrak{o}$ have embodied language $\Lang{\mathfrak{v}_\mathfrak{o}}$.
Let $INT_\mathfrak{o}\subseteq \Lang{\mathfrak{v}_\mathfrak{o}}$ be idealised self intervention episodes for $\mathfrak{o}$ and let $OBS_\mathfrak{o}\subseteq \Lang{\mathfrak{v}_\mathfrak{o}}$ be matched non self observations.
Any w-maximised causal identity
\[
\mathfrak{o}^1\in \mathcal{C}(INT_\mathfrak{o},OBS_\mathfrak{o})
\]
is called a \textbf{first order self} for $\mathfrak{o}$.
\end{definition}

\noindent\textbf{Interpretation.}
A first order self is the weakest self tag the organism can build that still separates self caused change from external change.

\begin{definition}[Chain notation for nested causal identity modelling]
Suppose we have two agents $\mathfrak{a}$ (Alice) and $\mathfrak{b}$ (Bob).
A causal identity $c^{\mathfrak{b}}_{\mathfrak{a}}$ denotes the internal model of $\mathfrak{b}$ constructed by $\mathfrak{a}$.

For a chain of agents $\mathfrak{x}_1,\dots,\mathfrak{x}_k$ we write
\[
 c^{\mathfrak{x}_1\mathfrak{x}_2\cdots \mathfrak{x}_k}_{\mathfrak{a}}
\]
for Alice's nested causal model in which $\mathfrak{x}_1$ models $\mathfrak{x}_2$ models $\dots$ models $\mathfrak{x}_k$.
For example, $c^{\mathfrak{ba}}_{\mathfrak{a}}$ denotes Alice's model of Bob's model of Alice.
\end{definition}

\noindent\textbf{Interpretation.}
This is just a notation for higher order theory of mind.

\begin{definition}[$n$th order self]
Fix an agent $\mathfrak{a}$.
An \textbf{$n$th order self} of $\mathfrak{a}$ is any causal identity whose chain ends at $\mathfrak{a}$ after interleaving $\mathfrak{a}$ with $n-1$ other agent models.
Formally set
\[
\mathfrak{a}^1 := c^{\mathfrak{a}}_{\mathfrak{a}}.
\]
For $n\ge 2$, choose agents $\mathfrak{x}_1,\dots,\mathfrak{x}_{n-1}$ and set
\[
\mathfrak{a}^n := c^{\mathfrak{x}_1\mathfrak{a}\mathfrak{x}_2\mathfrak{a}\cdots \mathfrak{x}_{n-1}\mathfrak{a}}_{\mathfrak{a}}.
\]
\end{definition}

\noindent\textbf{Interpretation.}
A second order self is a model of how others model you.
A third order self is a narrative self.
It is a model of how others model how you will act, and how that constrains your own future policy.
This is what makes long horizon identity, commitment, and trust possible.

\subsection{Organisms, protosymbols, and affect}

\begin{definition}[Organism]
We represent the circumstances of an organism $\mathfrak{o}$ as a tuple
\[
\mathfrak{o} = \langle \mathfrak{v}_\mathfrak{o}, \mu_\mathfrak{o}, \mathfrak{p}_\mathfrak{o}, <_\mathfrak{o} \rangle.
\]
Here $\mathfrak{v}_\mathfrak{o}$ is a vocabulary.
$\mu_\mathfrak{o}$ is a viability task that describes fit behaviour for $\mathfrak{o}$.
$\mathfrak{p}_\mathfrak{o}$ is a set of policies the organism is assumed to know.
$<_\mathfrak{o}$ is a preference ordering over tasks.
\end{definition}

\noindent\textbf{Interpretation.}
This is a compact way to say.
The organism has an embodied language.
It has a viability constraint.
It has limited memory and limited skills.
It has preferences.

\begin{definition}[Protosymbol system]
Let $\mathfrak{o}$ be an organism.
The \textbf{protosymbol system} of $\mathfrak{o}$ is
\[
\mathfrak{s}_\mathfrak{o} := \{\alpha \mid \exists\pi\in \mathfrak{p}_\mathfrak{o}\ \text{s.t.}\ \pi\in\Pi_\alpha\}.
\]
\end{definition}

\noindent\textbf{Interpretation.}
A protosymbol is a reusable meaning template.
It is a task the organism already knows how to solve.
The protosymbol system is the set of those templates.

\begin{definition}[Affect]
Let $\mathfrak{o}$ and $\mathfrak{a}$ be organisms.
We say $\mathfrak{o}$ affects $\mathfrak{a}$ when there exists a task $\alpha$ in $\mathfrak{s}_\mathfrak{a}$ and an input $i\in I_\alpha$ such that completing $\alpha$ from $i$ depends on what $\mathfrak{o}$ does.
\end{definition}

\noindent\textbf{Interpretation.}
Affect here is not a feeling word.
It means one organism can causally change the other organism's future options.

\section{Generalisation optimal learning}
\subsection{Maximally uninformative prior implies weakness maximisation}

We now give the formal result that justifies w-maxing as a generalisation optimal rule under a maximally uninformative prior.
This is the formal backbone of the Monte Carlo benchmark in the main text.

\begin{proposition}[Sufficiency of weakness maximisation under a uniform prior]
\label{prop:sufficiency}
Let $\mathfrak{v}$ be finite and let $\alpha=\langle I_\alpha,O_\alpha\rangle$ be a task in $\Lang{\mathfrak{v}}$.
Let
\[
U := \Lang{\mathfrak{v}}\setminus \Ext{I_\alpha}
\]
be the set of \emph{unseen} statements, meaning statements that never arise as completions of the child inputs.
Let $\omega$ be a parent task with $\alpha\sqsubseteq \omega$.
Assume a maximally uninformative extension model.
Conditional on $\alpha$, the parent selects an additional set $S\subseteq U$ of correct outputs uniformly at random from $2^U$, and sets
\[
O_\omega = O_\alpha \cup S.
\]
Then for any correct child policy $\pi\in\Pi_\alpha$,
\[
\mathbb{P}(\pi\in \Pi_\omega \mid \alpha) = \frac{2^{\card{\Ext{\pi}\cap U}}}{2^{\card{U}}}.
\]
In particular, among all $\pi\in\Pi_\alpha$, this generalisation probability is maximised by maximising $\card{\Ext{\pi}}$.
\end{proposition}

\begin{proof}
Fix a correct policy $\pi\in\Pi_\alpha$.
By correctness on $\alpha$,
\[
\Ext{I_\alpha}\cap \Ext{\pi} = O_\alpha.
\]
Because $U=\Lang{\mathfrak{v}}\setminus \Ext{I_\alpha}$, we can decompose
\[
\Ext{\pi} = (\Ext{\pi}\cap \Ext{I_\alpha})\ \cup\ (\Ext{\pi}\cap U) = O_\alpha\ \cup\ (\Ext{\pi}\cap U).
\]
Hence, within $\Pi_\alpha$, maximising $\card{\Ext{\pi}}$ is equivalent to maximising $\card{\Ext{\pi}\cap U}$. Under the stated random extension model, $\pi$ is also correct for the parent $\omega$ precisely when the additional correct outputs $S\subseteq U$ are all permitted by $\pi$.
That is when $S\subseteq \Ext{\pi}\cap U$.
The number of subsets $S\subseteq U$ satisfying $S\subseteq \Ext{\pi}\cap U$ is $2^{\card{\Ext{\pi}\cap U}}$.
The total number of subsets of $U$ is $2^{\card{U}}$.
Therefore
\[
\mathbb{P}(\pi\in\Pi_\omega\mid \alpha) = \frac{2^{\card{\Ext{\pi}\cap U}}}{2^{\card{U}}}.
\]
This quantity is strictly increasing in $\card{\Ext{\pi}\cap U}$.
So it is maximised by maximising $\card{\Ext{\pi}}$ within $\Pi_\alpha$.
\end{proof}

\noindent\textbf{Interpretation.}
We assume the unknown future cases are chosen uniformly from everything the training task did not rule out.
A policy survives the parent task if it already allows those future cases.
The more completions a policy allows, the more future cases it covers.
So weakness maximisation is Bayes optimal under this ignorance model.

\subsection{Weakness is necessary under the uniform prior}

\begin{proposition}[Necessity of weakness maximisation under a uniform prior]
\label{prop:necessity}
In the setting of Proposition~\ref{prop:sufficiency}, let $\pi_w\in\Pi_\alpha$ be a policy that maximises $\card{\Ext{\pi}}$ among correct child policies.
If $\pi$ is any other correct policy, then
\[
\mathbb{P}(\pi_w\in \Pi_\omega \mid \alpha) \ge \mathbb{P}(\pi\in \Pi_\omega \mid \alpha).
\]
If $\card{\Ext{\pi_w}} > \card{\Ext{\pi}}$ then the inequality is strict.
\end{proposition}

\begin{proof}
By Proposition~\ref{prop:sufficiency}, for any $\pi\in\Pi_\alpha$,
\[
\mathbb{P}(\pi\in\Pi_\omega\mid \alpha) = \frac{2^{\card{\Ext{\pi}\cap U}}}{2^{\card{U}}}.
\]
This is strictly increasing in $\card{\Ext{\pi}\cap U}$.
Within $\Pi_\alpha$, we have $\Ext{\pi} = O_\alpha \cup (\Ext{\pi}\cap U)$, so maximising $\card{\Ext{\pi}}$ is equivalent to maximising $\card{\Ext{\pi}\cap U}$.
Therefore any policy $\pi_w$ that maximises $\card{\Ext{\pi}}$ also maximises the generalisation probability.
If $\card{\Ext{\pi_w}} > \card{\Ext{\pi}}$, then $\card{\Ext{\pi_w}\cap U} > \card{\Ext{\pi}\cap U}$ and the inequality is strict.
\end{proof}

\noindent\textbf{Interpretation.}
If survival probability increases monotonically with weakness, then choosing the weakest correct policy is not just sufficient.
It is necessary.

\subsection{Weakness beyond the uniform prior}

The uniform prior from Proposition~\ref{prop:sufficiency} is the special case where every unseen statement becomes relevant with probability one half, independently.
A simple nonuniform generalisation keeps independence but allows different probabilities for different unseen statements.

\begin{theorem}[Bayes optimality under independent biased extension priors]
\label{thm:independent_prior}
Work in the setting of Proposition~\ref{prop:sufficiency} but replace the uniform prior on $S\subseteq U$.
Assume each $u\in U$ is included in $S$ independently with probability $r_u\in(0,1)$.
Let $A_\pi := \Ext{\pi}\cap U$.
Then for any $\pi\in\Pi_\alpha$,
\[
\mathbb{P}(\pi\in\Pi_\omega\mid \alpha) = \prod_{u\in U\setminus A_\pi} (1-r_u).
\]
Equivalently,
\[
\log \mathbb{P}(\pi\in\Pi_\omega\mid \alpha) = \sum_{u\in U\setminus A_\pi} \log(1-r_u) + \text{constant}.
\]
Therefore a Bayes optimal heuristic is obtained by maximising the weighted weakness score
\[
W_r(\pi):= \sum_{u\in A_\pi} w_u
\qquad\text{where}\qquad
w_u := -\log(1-r_u).
\]
When all $r_u$ are equal, $W_r(\pi)$ is proportional to $\card{A_\pi}$ and reduces to ordinary w-maxing.
\end{theorem}

\begin{proof}
By Proposition~\ref{prop:sufficiency}, $\pi\in\Pi_\omega$ iff $S\subseteq A_\pi$.
Under the independence assumption, $S\subseteq A_\pi$ holds iff every $u\in U\setminus A_\pi$ is excluded from $S$.
Each such exclusion occurs with probability $(1-r_u)$.
Independence gives the product formula.
Taking logs yields the second display.
Up to an additive constant independent of $\pi$, maximising this probability is the same as maximising $\sum_{u\in A_\pi} -\log(1-r_u)$.
\end{proof}

\noindent\textbf{Interpretation.}
If some unseen cases are more likely than others, the truly optimal rule is to keep open the cases that are likely to appear.
That is a weighted version of weakness.
Plain w-maxing is the special case where you really do not know which unseen case will matter.

\subsection{Why description length is not the right proxy}

\begin{proposition}[Simplicity can be suboptimal]
\label{prop:simp_suboptimal}
A shortest policy heuristic based on description length is neither sufficient nor necessary for maximising the generalisation probability in Proposition~\ref{prop:sufficiency}.
In particular, there exist tasks for which the shortest correct policy has strictly smaller weakness than another correct policy.
Under the uniform prior, the shortest correct policy is then strictly suboptimal.
\end{proposition}

\begin{proof}
We give an explicit counterexample. Let $\PhiEnv=\{\phi_0,\phi_1,\phi_2,\phi_3,\phi_4\}$.
Let the following five sets of programs be the maximal consistent statements.
\[
\begin{aligned}
 m_0&=\{z,j,k\},\\
 m_1&=\{j,k,a,b,c,d\},\\
 m_2&=\{j,k,e,f,g,h\},\\
 m_3&=\{u_j,j\},\\
 m_4&=\{u_k,k\}.
\end{aligned}
\]
Define the vocabulary $\mathfrak{v}:=\bigcup_{r=0}^4 m_r$.
Interpret each program $p\in\mathfrak{v}$ as the set of states in which it appears.
That is
\[
 p:=\{\phi_r\in\PhiEnv \mid p\in m_r\}.
\]
With this construction, a set of programs $\ell\subseteq \mathfrak{v}$ is a statement iff $\ell\subseteq m_r$ for some $r$.
So $\Lang{\mathfrak{v}}$ is exactly the downward closure of $\{m_0,\dots,m_4\}$. Now define a $\mathfrak{v}$-task $\alpha=\langle I_\alpha,O_\alpha\rangle$ by
\[
 I_\alpha=\bigl\{\{z,j,k\},\{u_j\},\{u_k\}\bigr\},
 \qquad
 O_\alpha=\bigl\{\{z,j,k\}\bigr\}.
\]
One can check that $\Ext{I_\alpha}=\{\{z,j,k\},\{u_j\},\{u_j,j\},\{u_k\},\{u_k,k\}\}$.\\

\noindent Consider two candidate policies.
Let $\pi_d:=\{z\}$ and $\pi_w:=\{j,k\}$.
Because the only element of $\Ext{I_\alpha}$ that contains $z$ is $\{z,j,k\}$, we have
$\Ext{I_\alpha}\cap \Ext{\pi_d}=\{\{z,j,k\}\}=O_\alpha$.
So $\pi_d\in\Pi_\alpha$.
Similarly, no element of $\Ext{I_\alpha}$ contains both $j$ and $k$ except $\{z,j,k\}$.
So $\Ext{I_\alpha}\cap \Ext{\pi_w}=O_\alpha$ and $\pi_w\in\Pi_\alpha$.\\

\noindent The length proxy prefers $\pi_d$ because $\card{\pi_d}=1<2=\card{\pi_w}$.
But weakness prefers $\pi_w$.
Program $z$ appears only in $m_0$, which gives $\card{\Ext{\pi_d}}=4$.
Programs $j$ and $k$ co-occur in $m_0,m_1,m_2$, which gives $\card{\Ext{\pi_w}}=32$.
So $\card{\Ext{\pi_w}} > \card{\Ext{\pi_d}}$.\\

\noindent Under Proposition~\ref{prop:sufficiency}, generalisation probability increases with $\card{\Ext{\pi}}$.
So the shortest policy $\pi_d$ is strictly suboptimal.
\end{proof}

\noindent\textbf{Interpretation.}
You can build vocabularies where the shortest description is a brittle hack.
It only works on the training cases.
A slightly longer policy can be much weaker in the important sense.
It covers more possible future cases.

\section{Full proofs for results used as sketches in the main text}
\subsection{Bayesian bound for the measurement failure mode}

This section fills in the proof sketch from the main text that links mismatch probability to a likelihood ratio.

\begin{proposition}[Likelihood ratio upper bound for decoder mismatch]
\label{prop:mismatch_upper}
Let an evaluator observe a trace $y$ and consider a hypothesis class $\mathcal{H}=\mathcal{H}_{mind}\cup\{h_{noise}\}$ where $h_{noise}$ treats the trace as noise.
Assume $\mathbb{P}(h_{noise})>0$ and $\mathbb{P}(y\mid h_{noise})>0$.
If there exists $\varepsilon\in(0,1)$ such that for every $h\in\mathcal{H}_{mind}$ we have
\[
\mathbb{P}(y\mid h)\le \varepsilon\,\mathbb{P}(y\mid h_{noise}),
\]
then
\[
\mathbb{P}(\mathcal{H}_{mind}\mid y)\le \frac{\varepsilon\,\mathbb{P}(\mathcal{H}_{mind})}{\varepsilon\,\mathbb{P}(\mathcal{H}_{mind})+\mathbb{P}(h_{noise})}.
\]
\end{proposition}

\begin{proof}
By Bayes rule,
\[
\mathbb{P}(\mathcal{H}_{mind}\mid y)=
\frac{\sum_{h\in\mathcal{H}_{mind}}\mathbb{P}(h)\mathbb{P}(y\mid h)}
{\sum_{h\in\mathcal{H}_{mind}}\mathbb{P}(h)\mathbb{P}(y\mid h)+\mathbb{P}(h_{noise})\mathbb{P}(y\mid h_{noise})}.
\]
Let $N=\sum_{h\in\mathcal{H}_{mind}}\mathbb{P}(h)\mathbb{P}(y\mid h)$ and $A=\mathbb{P}(h_{noise})\mathbb{P}(y\mid h_{noise})$.
Then $\mathbb{P}(\mathcal{H}_{mind}\mid y)=N/(N+A)$.
For every $h\in\mathcal{H}_{mind}$ we have $\mathbb{P}(y\mid h)\le \varepsilon\,\mathbb{P}(y\mid h_{noise})$, so
\[
N \le \sum_{h\in\mathcal{H}_{mind}}\mathbb{P}(h)\,\varepsilon\,\mathbb{P}(y\mid h_{noise})
= \varepsilon\,\mathbb{P}(\mathcal{H}_{mind})\,\mathbb{P}(y\mid h_{noise}).
\]
Substitute this into $N/(N+A)$ and cancel the common factor $\mathbb{P}(y\mid h_{noise})$.
\end{proof}

\noindent\textbf{Interpretation.}
If every mind hypothesis you know how to test fits the data much worse than ``it is just noise'', Bayes rule forces you to put almost all posterior mass on noise.
So a decoding mismatch can make a real mind look invisible.

\begin{proposition}[Likelihood ratio bound for mismatch]
\label{prop:mismatch}
Let $\theta$ denote the probability that an observation is drawn from the true process rather than a mismatched process.
Assume a Beta prior $\theta\sim\mathrm{Beta}(a,b)$.
Let $D$ be a dataset of $n$ observations and let $k$ of them match the true process.
Then the posterior is $\theta\mid D\sim \mathrm{Beta}(a+k,b+n-k)$.
The posterior odds that $\theta>1/2$ satisfy
\[
\log\frac{\mathbb{P}(\theta>1/2\mid D)}{\mathbb{P}(\theta\le 1/2\mid D)}
\ge
(a+k-b-n+k)\,\log 2.
\]
In particular, if $k$ is not substantially larger than $n/2$, the posterior cannot strongly favour $\theta>1/2$.
\end{proposition}

\begin{proof}
The Beta posterior is standard.
For the odds bound, write the posterior density as
\[
 f(\theta\mid D) \propto \theta^{a+k-1}(1-\theta)^{b+n-k-1}.
\]
Consider the likelihood ratio at $\theta$ versus $1-\theta$.
For $\theta\in(1/2,1)$,
\[
\frac{f(\theta\mid D)}{f(1-\theta\mid D)}
=\Bigl(\frac{\theta}{1-\theta}\Bigr)^{a+k-1-(b+n-k-1)}
=\Bigl(\frac{\theta}{1-\theta}\Bigr)^{a+k-b-n+k}.
\]
On $(1/2,1)$ we have $\theta/(1-\theta)\ge 1$.
So if $a+k-b-n+k\ge 0$, the posterior density is pointwise larger on $(1/2,1)$ than on $(0,1/2)$ after reflection, which yields the stated odds lower bound.
If $a+k-b-n+k<0$ the bound is still valid because the right hand side is then negative.
\end{proof}

\noindent\textbf{Interpretation.}
If you have not seen a strong signal that your decoder matches the world, you cannot justify treating your measurements as ground truth.
In that regime the right strategy is to probe.

\subsection{Causal identities are w-max separators}

\begin{proposition}[Existence of w-maximised causal identities in finite languages]
\label{prop:wmax_exists}
Assume $\mathfrak{v}$ is finite.
If $\mathcal{C}(INT,OBS)\neq\emptyset$, then a w-maximised causal identity exists.
\end{proposition}

\begin{proof}
When $\mathfrak{v}$ is finite, the language $\Lang{\mathfrak{v}}$ is finite.
So the candidate set $\mathcal{C}(INT,OBS)$ is finite.
The set of integers $\{\card{\Ext{c}}\mid c\in\mathcal{C}(INT,OBS)\}$ is therefore finite and nonempty.
So it has a maximum.
Any maximiser is a w-maximised causal identity.
\end{proof}

\noindent\textbf{Interpretation.}
If there are only finitely many signatures to check, the weakest one exists.

\begin{proposition}[Finite case w-max implies inclusion minimality]
\label{prop:wmax_inclusion_min}
Assume $\mathfrak{v}$ is finite.
Let $c^\star\in\mathcal{C}(INT,OBS)$ be w-maximised.
Then there is no other candidate $c\in\mathcal{C}(INT,OBS)$ such that $c\subset c^\star$.
\end{proposition}

\begin{proof}
Suppose for contradiction that $c\subset c^\star$ for some candidate $c$.
Then every completion of $c^\star$ is also a completion of $c$.
So $\Ext{c^\star}\subseteq \Ext{c}$.
Because $\mathfrak{v}$ is finite, $\Lang{\mathfrak{v}}$ is finite.
So strict inclusion implies strict cardinality.
Also, $c$ itself is in $\Ext{c}$ but not in $\Ext{c^\star}$.
So the inclusion is strict.
Hence $\card{\Ext{c^\star}}<\card{\Ext{c}}$.
That contradicts w-maximality.
\end{proof}

\noindent\textbf{Interpretation.}
If a weaker separator exists that still works, then the stronger one was never optimal.

\subsection{The first order self is a fixed point of w-maxing}

\begin{proposition}[Self fixed point]
\label{prop:self_fixed_point}
Let $\mathfrak{o}^1$ be a first order self as in Definition S1.10.
Then $\mathfrak{o}^1$ is a w-maximised causal identity for the self intervention discrimination problem.
Applying the w-max rule again returns a first order self.
\end{proposition}

\begin{proof}
By definition, $\mathfrak{o}^1$ is a w-maximised causal identity in the candidate set $\mathcal{C}(INT_\mathfrak{o},OBS_\mathfrak{o})$.
So it is already a fixed point of the selection rule that chooses w-maximised candidates.
\end{proof}

\noindent\textbf{Interpretation.}
The first self variable is defined as the weakest self tag that works.
So it is literally what you get by w-maxing.

\subsection{Second order selves and Gricean report}

This section formalises a simple but sharp dependency claim.
If what counts as a correct ``report'' depends on how an audience interprets a signal, then any signalling policy that is robust across audience variation must represent that audience variation.
In the main text, we interpret this representational requirement as a 2ND order self.

\begin{definition}[Audience dependent decoding]
Fix an intended meaning set $M$ and a signal set $\Sigma$.
Let $\Theta$ be a set of possible audience epistemic states.
For each $\theta\in\Theta$, let $D_\theta: \Sigma\to M$ be the decoding map used by an audience in epistemic state $\theta$.
\end{definition}

\noindent\textbf{Interpretation.}
Different audiences can decode the same signal differently.
We model that by allowing the decoder to depend on an audience state $\theta$.

\begin{definition}[Robust Gricean report]
A speaker chooses a signal by a policy $\pi$.
If the speaker intends meaning $m\in M$ and faces an audience in state $\theta\in\Theta$, the policy outputs a signal $s=\pi(m,\theta)$.
We say $\pi$ is \textbf{robustly correct} if for all $m\in M$ and all $\theta\in\Theta$ we have
\[
    D_\theta(\pi(m,\theta)) = m.
\]
We say $\pi$ \textbf{fails to model the audience} if $\pi(m,\theta)$ is independent of $\theta$.
\end{definition}

\noindent\textbf{Interpretation.}
Robust report means the listener recovers the intended meaning across audience variation.
Failing to model the audience means you emit the same signal for the same meaning no matter who is listening.

\begin{proposition}[Gricean report requires audience modelling]
\label{prop:grice_esm}
Assume there exist two audience states $\theta_1,\theta_2\in\Theta$ and some signal $s\in\Sigma$ such that $D_{\theta_1}(s)\neq D_{\theta_2}(s)$.
Then no robustly correct signalling policy can fail to model the audience.
Equivalently, any robustly correct policy must condition on a variable that distinguishes $\theta_1$ from $\theta_2$.
\end{proposition}

\begin{proof}
Suppose for contradiction that $\pi$ is robustly correct and independent of $\theta$.
Fix any meaning $m\in M$.
Because $\pi$ is independent of $\theta$, we have $\pi(m,\theta_1)=\pi(m,\theta_2)=:s$.
Robust correctness requires $D_{\theta_1}(s)=m$ and $D_{\theta_2}(s)=m$.
So $D_{\theta_1}(s)=D_{\theta_2}(s)$.
This contradicts the assumption that some signal has different decodings under $\theta_1$ and $\theta_2$.
Therefore any robustly correct policy must depend on $\theta$ through some audience modelling variable.
\end{proof}

\noindent\textbf{Interpretation.}
If two audiences interpret the same signal differently, you cannot talk to both of them correctly using one fixed rule that ignores who they are.
You need some representation of the audience.

\begin{corollary}[Gricean communication separates 1ST from 2ND order selves]
\label{cor:sep_grice_esm}
There exist communication tasks for which no policy can be robustly correct unless it conditions on an audience modelling variable.
In the main text's terminology, these are tasks that require a 2ND order self.
\end{corollary}

\begin{proof}
Pick any setting with $\Theta$ and $\{D_\theta\}_{\theta\in\Theta}$ that satisfies the assumption of Proposition~\ref{prop:grice_esm}.
By that proposition, robust correctness forces audience modelling.
So tasks of this form are solvable only if the agent's embodied language can represent the relevant audience distinctions.
\end{proof}

\noindent\textbf{Interpretation.}
Some kinds of report are impossible without modelling the audience.
That is a clean separation between having a self tag and having a socially usable report.

\subsection{Third order selves and self binding in trust games}

This section gives a minimal separation result that motivates 3RD order self structure.
It exhibits a multi stage social niche in which even perfect audience modelling is not enough.
A publicly legible self binding move is required to make trust rational.

\begin{definition}[A two stage trust and commitment game]
\label{def:trust_commit_esm}
Fix a commitment cost $c$ with $0<c<1$.
Two agents $\mathfrak{a}$ and $\mathfrak{b}$ play the following extensive form game.
\begin{enumerate}[leftmargin=*, itemsep=0.25em]
\item Stage 0. $\mathfrak{a}$ chooses either \emph{Bind} $B$ or \emph{Free} $F$.
Choosing $B$ incurs an immediate cost $c$ to $\mathfrak{a}$ and removes $\mathfrak{a}$'s ability to exploit in Stage 2.
Choosing $F$ incurs no cost and leaves Stage 2 unconstrained.
\item Stage 1. $\mathfrak{b}$ observes $B$ or $F$ and chooses either \emph{Trust} $T$ or \emph{Do not trust} $N$.
If $\mathfrak{b}$ chooses $N$, the game ends with payoff $(0,0)$.
\item Stage 2. If $\mathfrak{b}$ chose $T$ then the following happens.
\begin{itemize}[leftmargin=*, itemsep=0.15em]
\item if $\mathfrak{a}$ chose $B$ in Stage 0, $\mathfrak{a}$ is forced to play \emph{Honour} $H$.
\item if $\mathfrak{a}$ chose $F$ in Stage 0, $\mathfrak{a}$ chooses either \emph{Honour} $H$ or \emph{Exploit} $E$.
\end{itemize}
Payoffs, before subtracting the binding cost, are
\[
u_\mathfrak{a}(T,H)=1,\quad u_\mathfrak{b}(T,H)=1,\qquad
u_\mathfrak{a}(T,E)=2,\quad u_\mathfrak{b}(T,E)=-1.
\]
If $\mathfrak{a}$ chose $B$ in Stage 0, then $c$ is subtracted from $\mathfrak{a}$'s payoff.
\end{enumerate}
\end{definition}

\noindent\textbf{Interpretation.}
The listener decides whether to trust.
If they trust, the actor can either honour or exploit.
Binding is a move the actor can make in advance that removes its ability to exploit.

\begin{proposition}[A commitment game separates 2ND from 3RD order selves]
\label{prop:trust_commit_sep_esm}
In the game of Definition~\ref{def:trust_commit_esm}.
\begin{enumerate}[leftmargin=*, itemsep=0.25em]
\item If the self binding move $B$ is not available, the unique subgame perfect equilibrium outcome is $N$ with payoff $(0,0)$.
\item If $B$ is available and $0<c<1$, there exists a subgame perfect equilibrium in which $\mathfrak{a}$ plays $B$.
$\mathfrak{b}$ plays $T$ after observing $B$ and $N$ after observing $F$.
The Stage 2 outcome is necessarily $H$, yielding payoff $(1-c,1)$.
\end{enumerate}
\end{proposition}

\begin{proof}
If $B$ is unavailable, then after $\mathfrak{b}$ chooses $T$ the actor $\mathfrak{a}$ chooses between $H$ and $E$.
$\mathfrak{a}$ strictly prefers $E$ because $2>1$.
Anticipating this, $\mathfrak{b}$ strictly prefers $N$ to $T$ because $0>-1$.
So the unique subgame perfect equilibrium outcome is $N$ with payoff $(0,0)$. Now allow $B$.
In the subgame following $(F,T)$ the same reasoning applies.
$\mathfrak{a}$ plays $E$, so $\mathfrak{b}$'s best response after observing $F$ is $N$.
In the subgame following $(B,\cdot)$, $\mathfrak{a}$ is forced to play $H$, so $\mathfrak{b}$'s best response after observing $B$ is $T$ because $1>0$.
Anticipating these best responses, $\mathfrak{a}$ compares its Stage 0 payoffs.
Choosing $F$ yields $0$ since $\mathfrak{b}$ chooses $N$.
Choosing $B$ yields $1-c$.
Because $0<c<1$, we have $1-c>0$ so $\mathfrak{a}$ strictly prefers $B$.
This strategy profile is sequentially rational in every subgame.
So it is subgame perfect and yields the cooperative outcome $(1-c,1)$.
\end{proof}

\noindent\textbf{Interpretation.}
This is backward induction.
If the actor is unbound, exploitation is optimal and the listener anticipates that and refuses to trust.
If the actor binds, exploitation is unavailable so trust becomes rational.

\paragraph{Remark.}
In the main text we interpret a publicly legible self binding move as a minimal proxy for a 3RD order self operation.
It is a policy about policies.
Within Stack Theory, the same point can be expressed as a w maxing construction.
Among all self binding constraints that make trust viable, w maxing selects a maximally weak one.
It is strong enough to rule out exploitative completions, and weak enough to preserve flexibility everywhere else.

\section{Experiments and plots}
\subsection{Experiment S1: Weakness beats description length}
These experiments are taken from the workshop paper \cite{bennett2026b}. They are primarily illustrative, minimal demonstrations of the theoretical claims.

\paragraph{Goal.}
We test the claim that weakness, not description length, is the quantity that predicts generalisation under an ignorance prior.

\paragraph{World generation.}
For each world we sample.
A finite environment $\PhiEnv$.
A finite vocabulary $\mathfrak{v}\subseteq 2^{\PhiEnv}$.
A random child task $\alpha$ and a random parent task $\omega$ with $\alpha\sqsubseteq \omega$.
The parent is generated by adding extra outputs drawn from a prior over $2^U$.

\paragraph{Heuristics.}
A heuristic is a rule that chooses one policy from $\Pi_\alpha$.
We compare.
A w-max heuristic that chooses a policy with maximal weakness.
A shortest policy heuristic that chooses a policy with minimal description length in the given encoding.
A random heuristic that chooses uniformly from $\Pi_\alpha$.

\paragraph{Regret.}
Regret is measured in bits.
It is the expected negative log probability penalty for generalising to the sampled parent, relative to the Bayes optimal heuristic under the stated prior.

\begin{table}[h]
\centering
\caption{Experiment S1 summary. Mean regret in bits across worlds. Lower is better.}
\label{tab:exp1}
\begin{tabular}{lrr}
\toprule
Heuristic & Uniform prior & Nonuniform prior \\
\midrule
w-max & 0.0 & 0.225 \\
shortest & 13.4 & 7.27 \\
random & 70.2 & 63.5 \\
\bottomrule
\end{tabular}
\end{table}

\begin{figure}[h]
\centering
\includegraphics[width=0.85\linewidth]{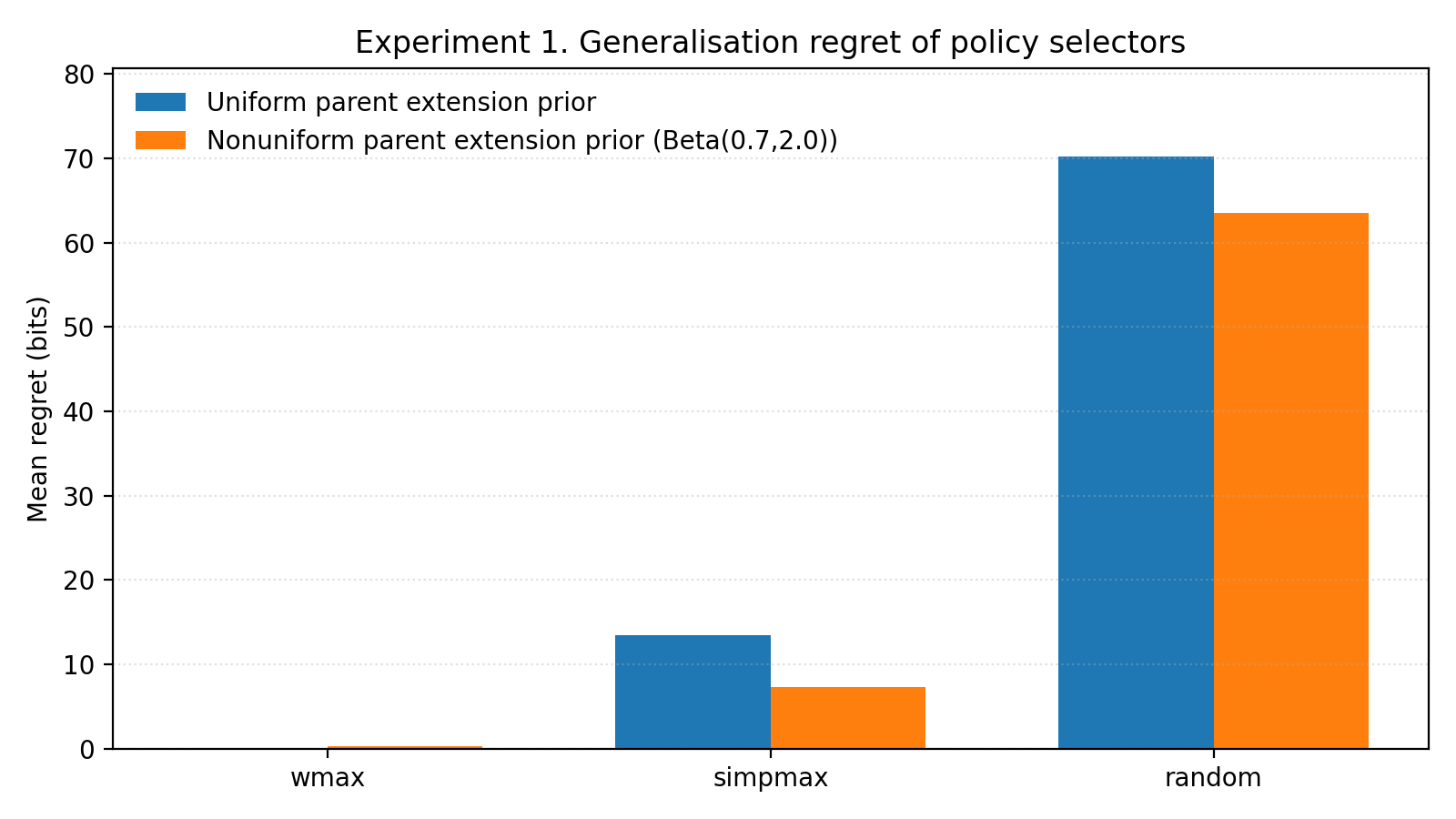}
\caption{Experiment S1. Regret in bits for three heuristics under two priors. w-max is optimal under the uniform prior by Proposition~\ref{prop:sufficiency}. It remains near optimal under the nonuniform prior.}
\label{fig:exp1}
\end{figure}

\subsection{Experiment S2: Probing under decoder mismatch}

\paragraph{Goal.}
We test the claim that under decoder mismatch, passive signalling is fragile while short probing closes most of the gap to an oracle.

\begin{table}[h]
\centering
\caption{Experiment S2 summary. Success rates for communication under mismatch.}
\label{tab:exp2}
\begin{tabular}{lrr}
\toprule
Method & Stationary decoder & Nonstationary decoder \\
\midrule
passive & 0.503 & 0.503 \\
probe $2$ & 0.926 & 0.928 \\
oracle & 0.961 & 0.961 \\
\bottomrule
\end{tabular}
\end{table}

\begin{figure}[h]
\centering
\includegraphics[width=0.85\linewidth]{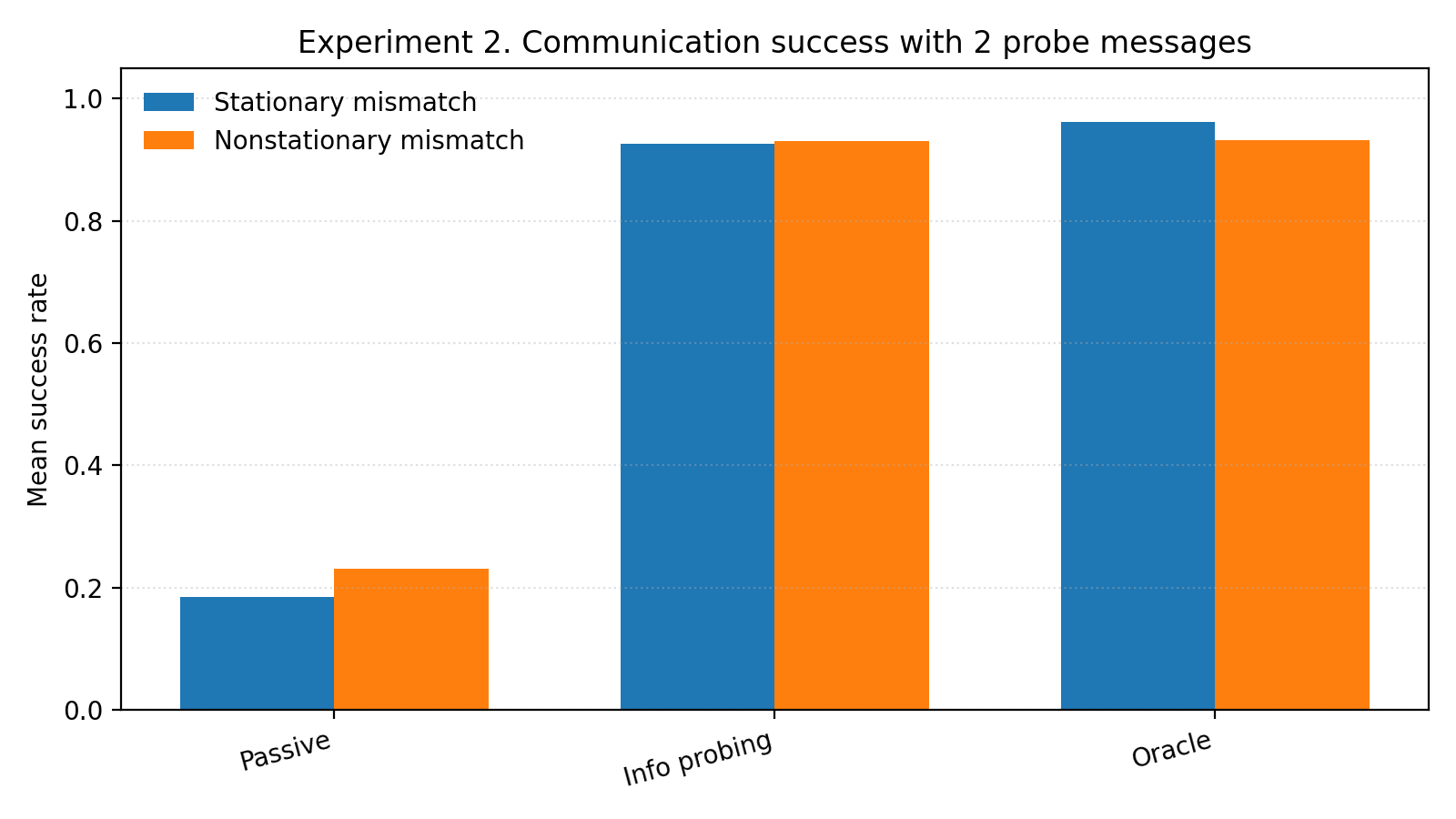}
\caption{Experiment S2. Passive signalling stays near chance. Two short probes achieve high success in both stationary and nonstationary settings.}
\label{fig:exp2}
\end{figure}

\subsection{Experiment S3: Commitment based trust}

\paragraph{Goal.}
We test the claim that credibility requires commitment.

\begin{table}[h]
\centering
\caption{Experiment S3 endpoints. Mean trust rates at horizon endpoints for binding and nonbinding commitments.}
\label{tab:exp3}
\begin{tabular}{lrr}
\toprule
Commitment type & $H=1$ & $H=5$ \\
\midrule
binding & 0.442 & 0.950 \\
nonbinding & 0.000 & 0.000 \\
\bottomrule
\end{tabular}
\end{table}

\begin{figure}[h]
\centering
\includegraphics[width=0.85\linewidth]{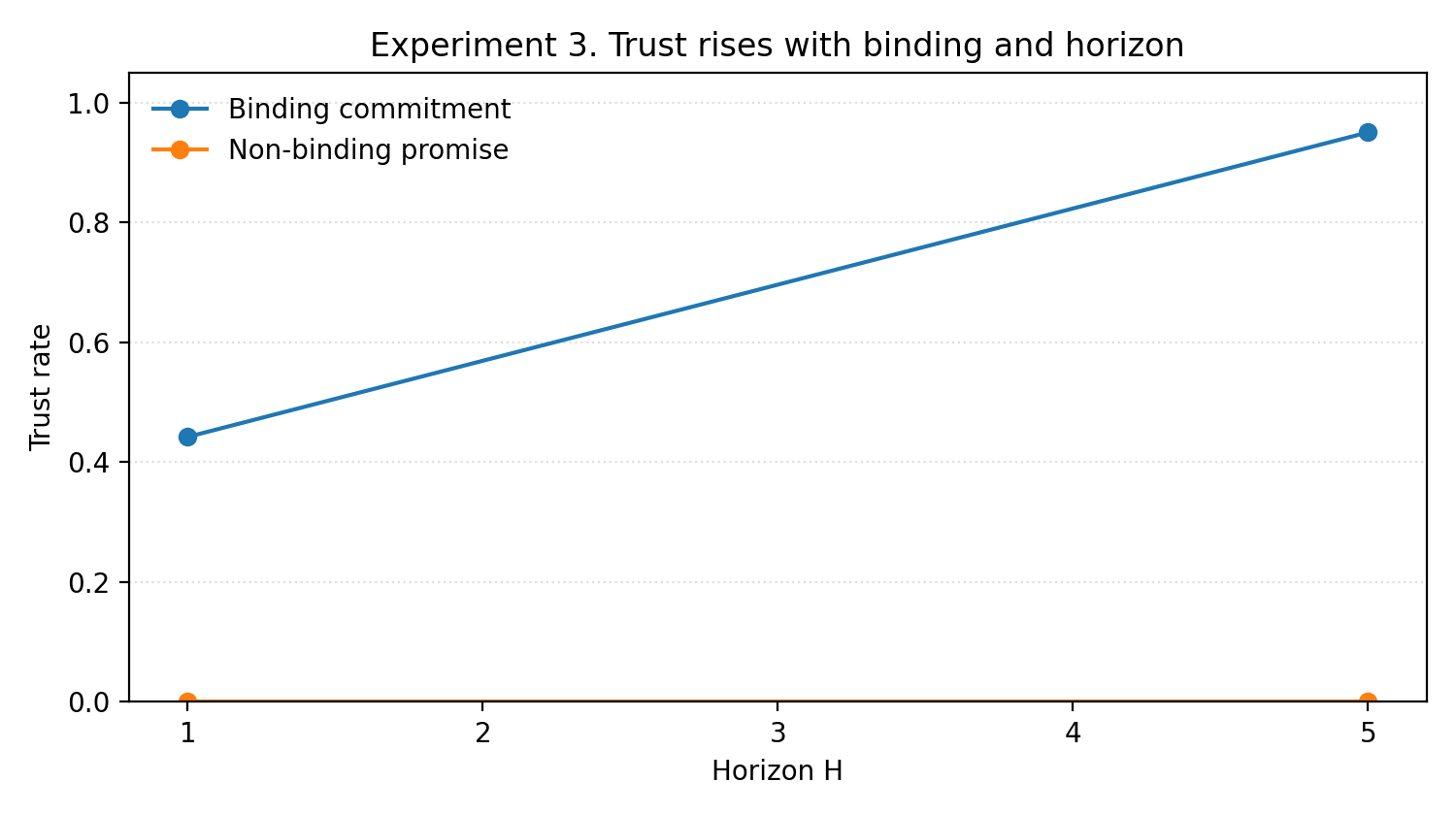}
\caption{Experiment S3. Trust rates at horizon endpoints. Binding dominates once the interaction has future.}
\label{fig:exp3a}
\end{figure}

\begin{figure}[h]
\centering
\includegraphics[width=0.85\linewidth]{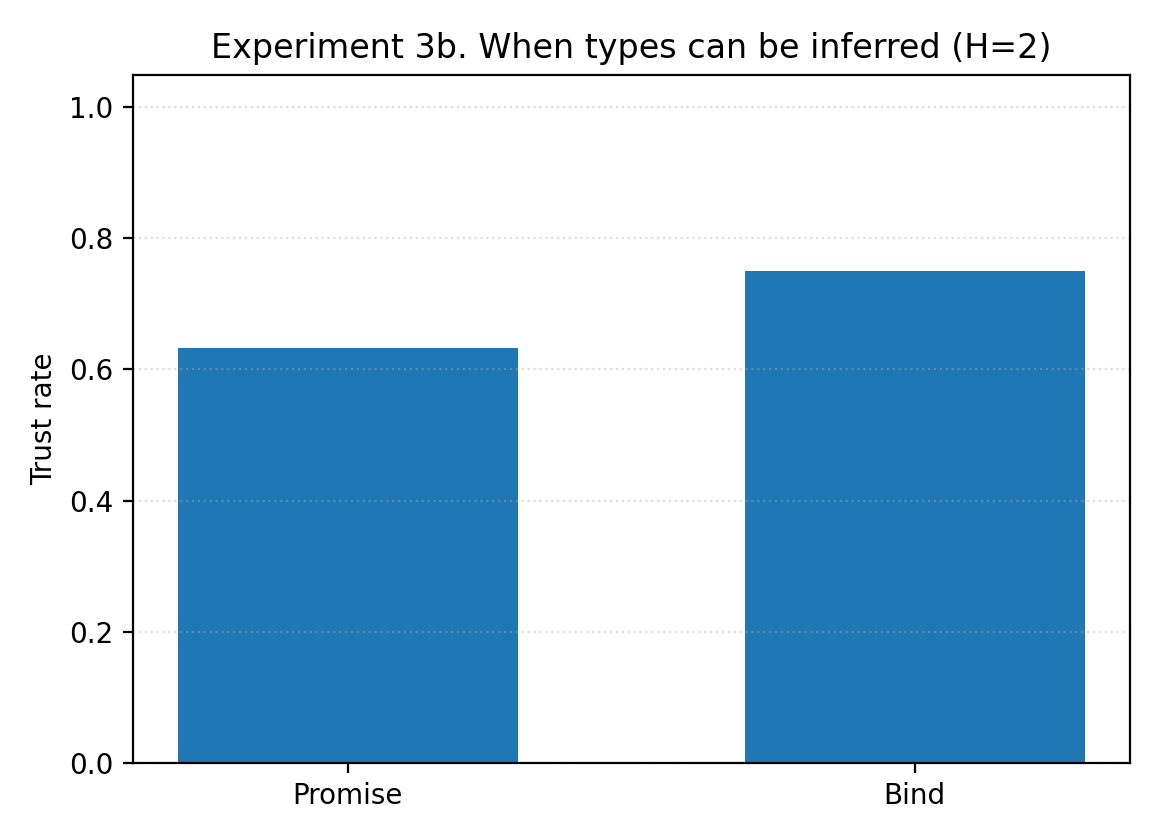}
\caption{Experiment S3. Distribution of game types in the benchmark sample.}
\label{fig:exp3b}
\end{figure}

\section{Law of the stack and a valence first causality principle}
\subsection{The Law of the Stack bottleneck}

The main text uses The Law of the Stack as a bottleneck result. The Law of the Stack was published in \cite{bennett2024c}.
It says higher level expressivity is exponentially limited by how many completions are left open below.

\begin{theorem}[The Law of the Stack]
\label{thm:lots}
Consider adjacent layers $i$ and $i+1$ and suppose
\[
\mathfrak{v}^{i+1} = \{\,\Truth(o) \mid o\in \Ext{\pi^i}\,\}.
\]
Let $\gamma^{i+1}$ be the induced layer $i+1$ task.
Then
\[
\sup_{\pi\in\Pi_{\gamma^{i+1}}}\card{\Ext{\pi}} \le 2^{\card{\Ext{\pi^i}}}.
\]
If $\card{\Ext{\pi^i}}<\infty$ then the finite bottleneck form holds
\[
\epsilon(\gamma^{i+1}) + \card{O_{\gamma^{i+1}}} \le 2^{\card{\Ext{\pi^i}}}.
\]
\end{theorem}

\begin{proof}
By definition,
\[
\mathfrak{v}^{i+1}=\{\,\Truth(o)\mid o\in \Ext{\pi^i}\,\}.
\]
So $\mathfrak{v}^{i+1}$ is the image of $\Ext{\pi^i}$ under $o\mapsto\Truth(o)$.
Therefore $\card{\mathfrak{v}^{i+1}}\le \card{\Ext{\pi^i}}$. Every statement in $\Lang{\mathfrak{v}^{i+1}}$ is a subset of $\mathfrak{v}^{i+1}$.
So $\Lang{\mathfrak{v}^{i+1}}\subseteq 2^{\mathfrak{v}^{i+1}}$.
Hence
\[
\card{\Lang{\mathfrak{v}^{i+1}}} \le 2^{\card{\mathfrak{v}^{i+1}}} \le 2^{\card{\Ext{\pi^i}}}.
\]
Now fix any $\pi\in\Pi_{\gamma^{i+1}}\subseteq \Lang{\mathfrak{v}^{i+1}}$.
By definition of extension, $\Ext{\pi}\subseteq \Lang{\mathfrak{v}^{i+1}}$.
So
\[
\card{\Ext{\pi}}\le \card{\Lang{\mathfrak{v}^{i+1}}}\le 2^{\card{\Ext{\pi^i}}}.
\]
Taking the supremum over $\pi\in\Pi_{\gamma^{i+1}}$ gives the first claim. For the finite bottleneck form, assume $\card{\Ext{\pi^i}}<\infty$.
Then $\mathfrak{v}^{i+1}$ and $\Lang{\mathfrak{v}^{i+1}}$ are finite.
So the supremum in the definition of utility is a maximum.
Choose $\pi^\star\in\Pi_{\gamma^{i+1}}$ that attains it.
Because $\pi^\star$ is correct, we have $O_{\gamma^{i+1}}\subseteq \Ext{\pi^\star}$.
All sets are finite, so
\[
\card{\Ext{\pi^\star}} = \card{O_{\gamma^{i+1}}} + \card{\Ext{\pi^\star}\setminus O_{\gamma^{i+1}}} = \card{O_{\gamma^{i+1}}} + \epsilon(\gamma^{i+1}).
\]
Also $\card{\Ext{\pi^\star}}\le 2^{\card{\Ext{\pi^i}}}$ by the first part.
Rearranging gives the second claim.
\end{proof}

\noindent\textbf{Interpretation.}
If the lower layer leaves only a small number of consistent refinements open, then the next layer can only name a limited number of distinct things.
It is a combinatorial bottleneck.

\subsection{Free energy floor from hard commitments}

We can connect the bottleneck to a simple free energy proxy.

\begin{corollary}[Free energy floor]
\label{cor:fep}
Assume the finite setting of Theorem~\ref{thm:lots}.
Fix a layer $i+1$ viability statement $\mu\in \Lang{\mathfrak{v}^{i+1}}$.
For any policy $\pi\in\Lang{\mathfrak{v}^{i+1}}$ define
\[
\Omega_\pi := \Ext{\pi}\cap \Ext{\mu}.
\]
Assume $\Omega_\pi\neq\emptyset$.
Let $q_\pi$ be the uniform distribution on $\Omega_\pi$ and let $p_\mu$ be the uniform distribution on $\Ext{\mu}$.
Define the base 2 free energy proxy
\[
F_2(q_\pi) := D_{\mathrm{KL},2}(q_\pi\,\|\,p_\mu).
\]
Then
\[
F_2(q_\pi)=\log_2\card{\Ext{\mu}}-\log_2\card{\Omega_\pi}.
\]
Also $\card{\Omega_\pi}\le 2^{\card{\Ext{\pi^i}}}$.
So
\[
F_2(q_\pi)\ge \log_2\card{\Ext{\mu}} - \card{\Ext{\pi^i}}.
\]
\end{corollary}

\begin{proof}
The KL identity for uniform distributions is direct.
If $q$ is uniform on a set of size $m$ and $p$ is uniform on a set of size $n$ containing that set, then $D_{\mathrm{KL},2}(q\|p)=\log_2(n/m)$.
Here $n=\card{\Ext{\mu}}$ and $m=\card{\Omega_\pi}$.

For the bound, we have $\Omega_\pi\subseteq \Lang{\mathfrak{v}^{i+1}}$.
Also $\Lang{\mathfrak{v}^{i+1}}\subseteq 2^{\mathfrak{v}^{i+1}}$.
So $\card{\Omega_\pi}\le 2^{\card{\mathfrak{v}^{i+1}}}$.
By Theorem~\ref{thm:lots} we have $\card{\mathfrak{v}^{i+1}}\le \card{\Ext{\pi^i}}$.
Combine these inequalities.
\end{proof}

\noindent\textbf{Interpretation.}
If you over constrain the lower layer, the number of viable futures at the next layer is capped.
That puts a floor under free energy.

\subsection{A valence first psychophysical causality principle}

We can now state the proof idea requested in the main text.
Generalisation optimal learning pushes the primitive ontology toward valence.
Anything else is a hard commitment that shrinks the viable future set and raises the free energy floor.

\begin{definition}[Hard primitive commitments]
Fix a language $\Lang{\mathfrak{v}}$.
Let $C\subseteq \Lang{\mathfrak{v}}$ be a set of commitments.
We say a learning rule has \textbf{hard primitives} $C$ if it restricts the policy space to policies that include every commitment in $C$.
For a task $\alpha$ this restricted class is
\[
\Pi^{C}_\alpha := \{\pi\in\Pi_\alpha \mid \forall c\in C,\ c\subseteq \pi\}.
\]
\end{definition}

\noindent\textbf{Interpretation.}
Hard primitives are constraints you force every candidate policy to contain.
They are not learned.
They are assumed in advance.

\begin{theorem}[The Psychophysical Principle of Causality]
\label{thm:valence_first}
Valence precedes representation under generalisation optimality.
Fix a viability statement $\mu\in \Lang{\mathfrak{v}}$.
For any policy $\pi\in \Lang{\mathfrak{v}}$ define its viable continuation set
\[
\Omega_\pi := \Ext{\pi}\cap \Ext{\mu}.
\]
Let $q_\pi$ be the uniform distribution on $\Omega_\pi$ and let $p_\mu$ be the uniform distribution on $\Ext{\mu}$.
When $\Omega_\pi\neq\emptyset$, define the base 2 free energy proxy
\[
F_2(\pi) := D_{\mathrm{KL},2}(q_\pi\,\|\,p_\mu) = \log_2\card{\Ext{\mu}}-\log_2\card{\Omega_\pi}.
\]

\noindent Let $C\subseteq \Lang{\mathfrak{v}}$ be a nonempty set of hard primitive commitments.
Assume $C$ contains at least one commitment $c$ that is not forced by viability.
That means there exists some viable continuation $o\in \Ext{\mu}$ with $c\not\subseteq o$.\\

\noindent Let $\pi$ be any policy with $\Omega_\pi\neq\emptyset$.
Assume also that $\Omega_\pi$ contains at least one viable continuation that violates $c$.
Then either $\Omega_{\pi\cup c}=\emptyset$ or
\[
F_2(\pi\cup c) > F_2(\pi).
\]
So adding an unnecessary primitive commitment strictly raises free energy whenever it rules out even one viable future. In particular, any learning rule that enforces non valence representational primitives as hard coded commitments pays a strict free energy penalty on tasks where the evidence does not already force those commitments.
So under a generalisation optimality objective, valence is the only safe primitive.
Everything else must be learned as a revisable causal identity.
\end{theorem}

\begin{proof}
Fix $c$ as in the statement.
We have $\Ext{\pi\cup c}\subseteq \Ext{\pi}$ because every completion of $\pi\cup c$ is a completion of $\pi$.
Intersect with viability to get $\Omega_{\pi\cup c}\subseteq \Omega_\pi$.\\

\noindent If $\Omega_{\pi\cup c}=\emptyset$ we are done.
Otherwise, by assumption there exists $o\in \Omega_\pi$ with $c\not\subseteq o$.
Then $o\notin \Omega_{\pi\cup c}$.
So the inclusion is strict and $\card{\Omega_{\pi\cup c}}<\card{\Omega_\pi}$.\\

\noindent Apply the identity $F_2(\pi)=\log_2\card{\Ext{\mu}}-\log_2\card{\Omega_\pi}$.
Since the numerator is fixed and the denominator shrinks, $F_2(\pi\cup c)>F_2(\pi)$.
\end{proof}

\noindent\textbf{Interpretation.}
If you hard code extra ontology, you eliminate viable futures.
That puts a floor under free energy and it blocks generalisation optimality.
So the only thing you are allowed to treat as primitive is the viability signal itself.
That is what we call valence.
Everything else has to be learned as a revisable causal identity.

\bibliographystyle{plainnat}
\bibliography{master_bibliography}